\documentclass[journal]{IEEEtran}
\usepackage{graphicx}
\usepackage{soul}
\usepackage[hyphens]{url} 
\usepackage{hyperref}
\usepackage{pifont}
\usepackage{verbatim}
\usepackage{subfigure}
\usepackage{calc}
\usepackage{amstext,amsmath}
\usepackage{stackengine}
\newcommand\IncG[2][]{\addstackgap{%
\raisebox{-.5\height}{\includegraphics[#1]{#2}}}}
\usepackage{array}
\usepackage{amsmath,amssymb} 
\usepackage{xcolor}
\usepackage{paralist}
\usepackage{multirow}
\usepackage{hyperref}
\usepackage{algorithm}  
\usepackage{algpseudocode}  
\renewcommand{\algorithmicrequire}{\textbf{Input:}}

\usepackage{anyfontsize}
\usepackage{eqparbox}
\newdimen{\algindent}
\algnewcommand\LeftComment[2]{%
\hspace{#1\algindent}$\triangleright$ \eqparbox{COMMENT}{#2} \hfill %
}

\ifCLASSINFOpdf
\else
\fi
\begin{document}
%
\title{Progressive Self-Distillation for Ground-to-Aerial Perception Knowledge Transfer}
%
%
%

\author{Junjie Hu,~\IEEEmembership{Member,~IEEE,}
         Chenyou Fan,
         Mete Ozay,
         Hua Feng,
         Yuan Gao,
         and Tin Lun Lam,~\IEEEmembership{Senior~Member,~IEEE}
\IEEEcompsocitemizethanks{\IEEEcompsocthanksitem J.Hu, H.Feng, Y.Gao and T.L.Lam are with the Shenzhen Institute of Artificial Intelligence and Robotics for Society (AIRs), Shenzhen, China.
E-mail: hujunjie@cuhk.edu.cn, fenghua0thomershen@gmail.com, gaoyuan@cuhk.edu.cn.
\IEEEcompsocthanksitem C.Fan is with the School of Artificial Intelligence, South China Normal University, China. E-mail: fanchenyou@scnu.edu.cn.
\IEEEcompsocthanksitem M.Ozay is with the Samsung Research, UK.  E-mail: meteozay@gmail.com.
\IEEEcompsocthanksitem T.L.Lam is also with the School of Science and Engineering, the Chinese University of Hong Kong, Shenzhen, China.
E-mail:tllam@cuhk.edu.cn.
\IEEEcompsocthanksitem T.L.Lam is the corresponding author.}
}

\maketitle

\begin{abstract}
We study a practical yet hasn't been explored problem: how a drone can perceive in an environment from different flight heights. Unlike autonomous driving, where the perception is always conducted from a ground viewpoint, a flying drone may flexibly change its flight height due to specific tasks, requiring the capability for viewpoint invariant perception. Tackling the such problem with supervised learning would entail tremendous costs for data annotation of different flying heights. On the other hand, current semi-supervised learning methods are not effective under viewpoint differences. In this paper, we introduce the ground-to-aerial perception knowledge transfer and propose a progressive semi-supervised learning framework that enables drone perception using only labeled data of ground viewpoint and unlabeled data of flying viewpoints. Our framework has four core components: i) a dense viewpoint sampling strategy that splits the range of vertical flight height into a set of small pieces with evenly-distributed intervals, ii)  nearest neighbor pseudo-labeling that infers labels of the nearest neighbor viewpoint with a model learned on the preceding viewpoint, iii) MixView that generates augmented images among different viewpoints to alleviate viewpoint differences, and iv) a progressive distillation strategy to gradually learn until reaching the maximum flying height. We collect a synthesized and a real-world dataset, and we perform extensive experimental analyses to show that our method yields $25.7\%$ and $16.9\%$ relative accuracy improvement for the synthesized dataset and the real world. Code and datasets are available on \url{https://github.com/FreeformRobotics/Progressive-Self-Distillation-for-Ground-to-Aerial-Perception-Knowledge-Transfer}.
\end{abstract}

\IEEEpeerreviewmaketitle

\section{Introduction}

\IEEEPARstart{R}OBOT perception plays a critical role in understanding interactive environments and providing substantial instruments for subsequent task execution. 
To date, most autonomous perception systems have focused on unmanned ground vehicles (UGVs) \cite{RobotCarDatasetIJRR,Zurn2021SelfSupervisedVT,Geiger2012CVPR}, e.g. autonomous driving cars. However, only a small portion of works studied drone perception due to hardness of the task and data collection.

Modern paradigm of machine learning introduces a data-driven deep learning based solution to address perception problems which has been validated its effectiveness on various tasks, e.g., semantic segmentation \cite{noh2015learning,long2015fully}, object detection \cite{redmon2016you,ceola2022learn},  depth regression \cite{Hu2019RevisitingSI,Hu2021BoostingLD,SunYXLSS22}, classification \cite{He2016DeepRL,huang2017densely}, and therefore provides a potential solution to autonomous driving. Recent studies on drone perception \cite{zhu2018visdrone,zhu2019visdrone,cao2021visdrone} have attempted to transfer this paradigm from cars to drones, i.e., first to prepare a labeled training set of drones and then to learn a deep network on a specific task.

\begin{figure}[t]
\centering
\includegraphics[width=0.48\textwidth]{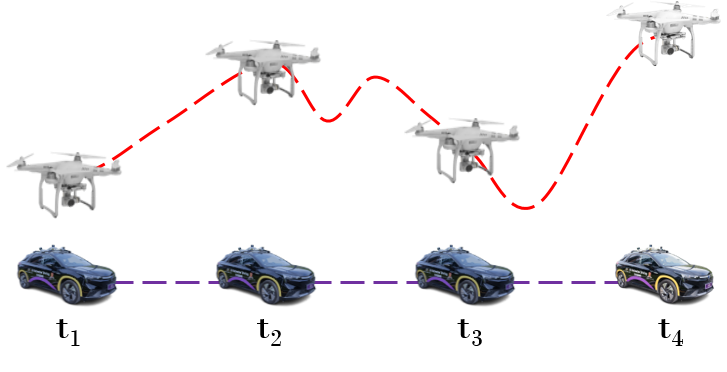}
\vspace{-5mm}
\caption{An example of a trajectory comparison between a drone and a car. One essential difference is that a drone is capable of optionally changing its flight height at time instance $\mathbf{t}_j, j=1,2,3,4$.}
\label{fig_trajectory}
\end{figure}

On the other hand, prior works scarcely considered the essential difference of the perception between drones and cars. 
In many works \cite{Gawel2018XViewGS,Guo2021SemanticHB}, the perception is assumed to be conducted at only a certain flying height.
However, in practical scenarios, as shown in Fig.~\ref{fig_trajectory}, a drone may appear at an arbitrary flight height due to its specifically allocated task. {\it Given the maximum flight height $h$, we argue that an essential requirement of drone perception is the capability to precisely perceive from different flight heights in the range of $[0, h]$ meters.}

It is, however, hard to tackle the above problem in a fully supervised learning manner. We may need to prepare even dozens of times more training data to obtain  performance comparable to that of an autonomous driving car. In reality,
data acquisition of ground truths is both costly and time-consuming. For instance, robot vision tasks, including object detection, semantic segmentation, and classification, require annotation of RGB images to be manual labeling, which is an unavoidable obstacle for developing data-driven approaches.

\begin{figure}[t]
\centering
\includegraphics[width=0.45\textwidth]{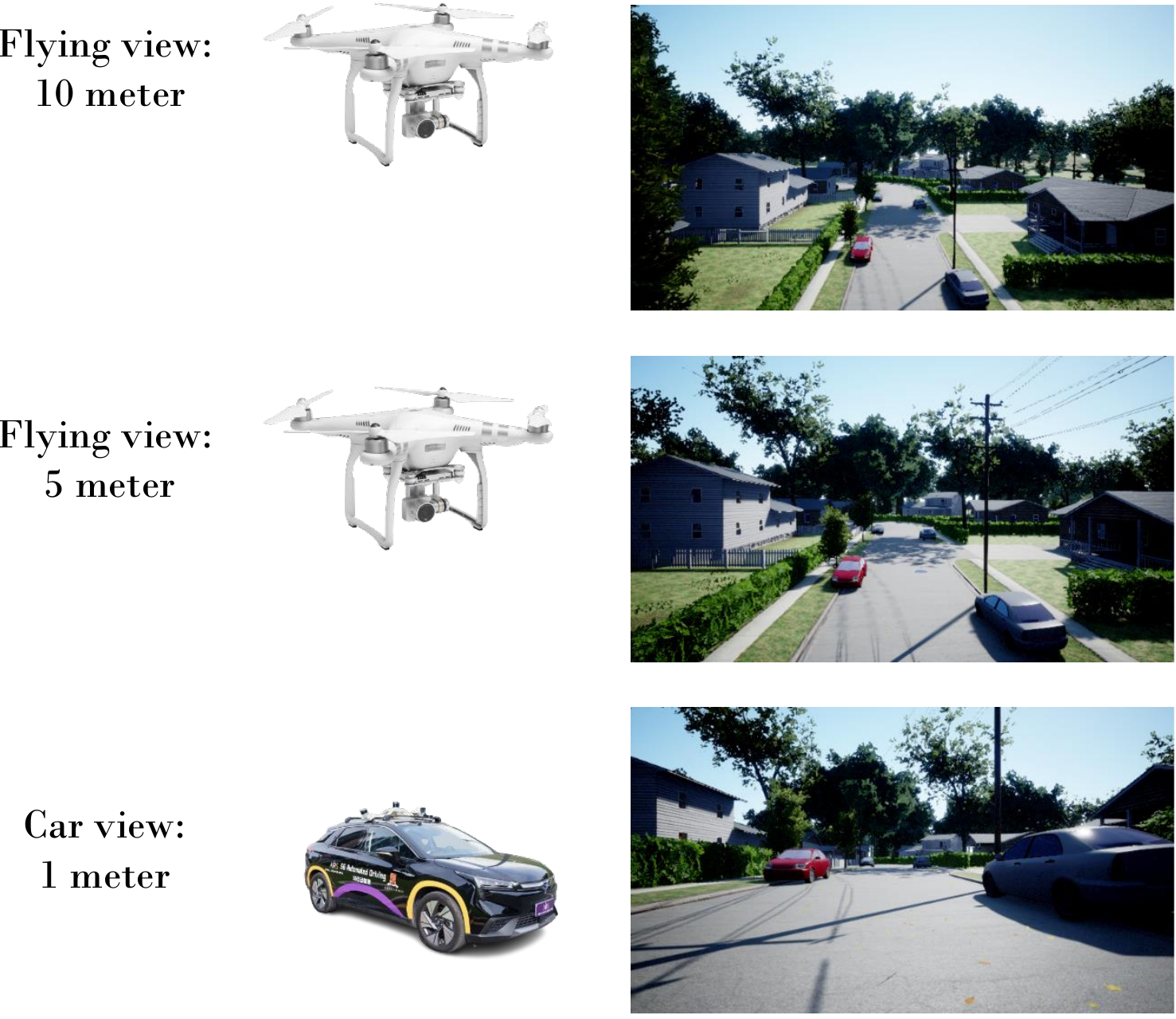}
\caption{Examples of  RGB images captured in a simulated environment at different heights.}
\label{fig_example}
\end{figure}

In this paper, we introduce the concept of ground-to-aerial (GoA) perception knowledge transfer that transfers the perception knowledge from a UGV for ground viewpoint perception to our drone perception tasks, without additional data annotation at different flying heights other than ground viewpoint.
To this end, we consider a semi-supervised learning (SSL) approach to enable drone perception where only data collected from the ground viewpoint are labeled, and all images captured from flying viewpoints are unlabeled.
The fundamental challenge is the inaccuracy caused by viewpoint differences among different flight heights. Fig.~\ref{fig_example} shows several images captured at three vertically different viewpoints, and they demonstrate clear viewpoint differences. 

To overcome the above difficulty, we first propose a dense viewpoint sampling strategy that splits the vertical flying range to heights with evenly-distributed intervals, and at each height, we sample data from that viewpoint. Even though the performance gap of large viewpoint differences is significant, we observe that the nearest viewpoint yields similar performance. Hence, we propose the nearest neighbor pseudo-labeling that predicts data of the viewpoint $h_i$ with a network learned on its nearest preceding viewpoint $h_{i-1}$. In addition, we propose MixView which mixes data from different viewpoints to generate augmented images, which is beneficial for relieving the issue of viewpoint change among data samples. We further propose a progressive SSL framework to gradually learn a model from the ground viewpoint until reaching the maximum flight height.

To facilitate this work, we create both a synthesized dataset collected from AirSim \cite{airsim2017fsr} and a real-world dataset captured from a city street. Both datasets include images of different flying heights. Note that, in this paper, we technically consider the vertical viewpoint difference which is exclusive for drones, rather than the horizontal viewpoint difference.
We study drone perception on semantic segmentation which is a fundamental task of robot perception.
We evaluate our method on those two datasets and provide both quantitative and qualitative analyses.

In summary, our contributions include:
\begin{itemize}
    \item To the best of our knowledge, the present study is the first attempt to tackle drone perception under large viewpoint change caused by different flight heights. 
   We aim at enabling drone perception to be precise from different flight heights.

    \item A novel semi-supervised learning framework that solves ground-to-aerial perception knowledge distillation with labeled images captured from the ground viewpoint with unlabeled images captured from flight viewpoints.
    
    \item Two proposed methods, i) the nearest neighbor pseudo-labeling and ii) MixView, to overcome the significant viewpoint change among different flying heights.

    \item Two elaborately crafted datasets for drone perception on semantic segmentation, including i) a synthesized one collected from AirSim with fixed lighting conditions and objects, and ii) a real-world one taken from a city street with changing lighting conditions and dynamic objects. 

\end{itemize}

The remainder of this paper is organized as follows. In Sec.~\ref{sec_related_work}, we discuss the necessary background and related studies. We describe our progressive semi-supervised learning framework in Sec.~\ref{sec_method}.  We then present two self-collected datasets used to verify our method in Sec.~\ref{sec_dataset}. Finally, we provide extensive numerical evaluations in Sec.~\ref{sec_result}.

\section{Related Work}
\label{sec_related_work}
\subsection{Drone Perception}
Consumer drones are usually equipped with only optical cameras, and thus enabling successful robotic perception from RGB images captured by these cameras will greatly boost the further development of autonomous drones. 
Similar to modern studies on self-driving cars, prior works tackle it as a classic data-driven learning problem, i.e., collecting a set of training data and training a deep neural network for perception. Early  works addressed indoor gate detection for drone navigation with a convolutional neural network \cite{Jung2018PerceptionGA}, or aimed at directly learning navigation in GPS-denied indoor corridor arenas \cite{Padhy2018DeepNN}.
The same tendency is also observed in object detection \cite{Vaddi2019EfficientOD,Mittal2020DeepLO,Zhu2021DetectionAT}, and semantic segmentation \cite{LYU2020108,Girisha2019SemanticSO}.

One fundamental difference of drone perception from autonomous driving is that a drone may fly at an arbitrary height, while self-driving cars only perform perception from a ground viewpoint. 
Drone perception is, therefore, more challenging, as it has to be accurate from different flying heights. A straightforward solution is to label all data from different viewpoints, and solve it in a purely supervised learning setup. However, it is not practical as all efforts will be put into data labeling. Therefore, in this paper, we propose a semi-supervised approach that enables drone perception under large viewpoint change.

\begin{figure*}[!t]
\centering
\includegraphics[width=0.95\textwidth]{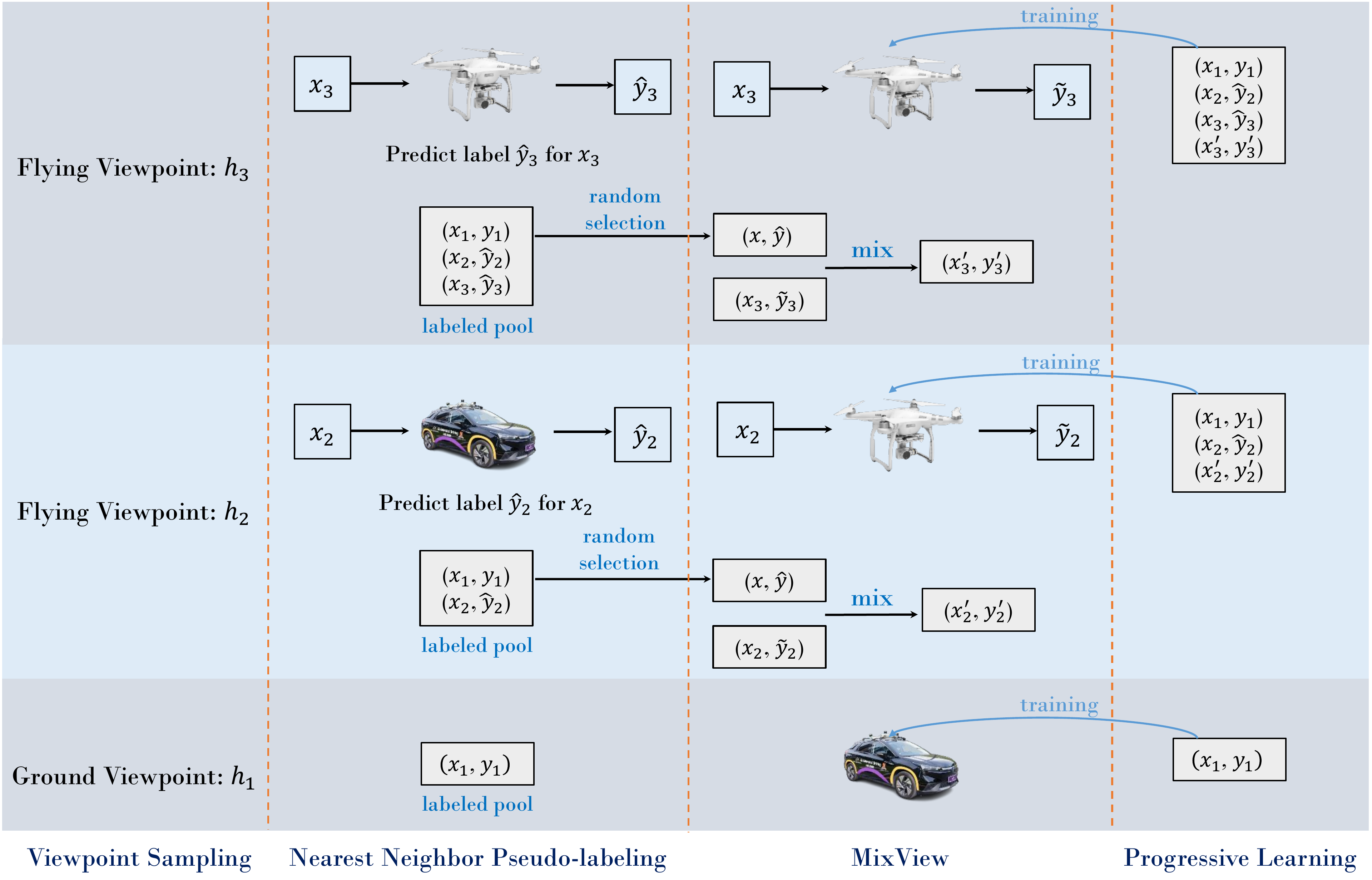}
\caption{The diagram of our progressive semi-supervised learning framework. We propose to transfer the knowledge from a UGV to enable drone perception while assuming only data from the ground viewpoint is labeled and all data from other viewpoints are not labeled. 
Here, we demonstrate an application scenario of three viewpoints, i.e., one ground viewpoint $h_1$ and two flying viewpoints $h_2$ and $h_3$, and $h_3 > h_2 > h_1$ in flight height.  $x_1, x_2, x_3$ are images taken from each viewpoint, respectively. $y_1$ is label of $x_1$, and $(x_1, y_1)$ are employed to train the UGV. To enable drone perception at $h_2$, we propose the nearest neighbor pseudo-labeling to infer labels of $x_2$ with the UGV. Moreover, we propose MixView to mix data of different viewpoints to generate augmented samples. The labeled samples, as well as the augmented samples, are then used to learn at $h_2$. The same learning strategy is then utilized to learn at $h_3$.
}
\label{fig_method}
\end{figure*}

\subsection{Multi-Robot Perception}
Several prior works studied homogeneous multi-robot collaborative semantic segmentation. 
In \cite{Liu2020When2comMP,Liu2020Who2comCP}, online collaboration methods were proposed to boost the performance of a single robot when its captured images suffer from deterioration.
Recently, a few-shot learning based approach \cite{Fan2021FewShotMP} was proposed to solve multi-robot perception in data-scarce scenarios.
However, the viewpoint difference between homogeneous robots is assumed to be small in these methods.

In this work, we introduce the ground-to-aerial perception distillation. To the best of our knowledge, we are the first to demonstrate the possibility of drone perception by purely learning from a UGV. Unlike \cite{Liu2020Who2comCP, Liu2020When2comMP, Fan2021FewShotMP}, our method can be applied to a heterogeneous multi-robot system consisting of a UGV and a UAV for perception in fully 3D spaces. Besides, unlike prior works on multi-robot systems \cite{xin2021multimobile,li2021memetic} that assume multiple sensors are available, we target more hard cases where only a visual camera is mounted on a UAV.

\subsection{Semi-supervised Learning}
Semi-supervised learning (SSL) is an active research field and has been extensively studied in machine learning. It seeks to learn a highly discriminative model from both labeled and unlabeled samples
\cite{Yang2021ASO,vanEngelen2019ASO}.
Prior studies have shown promising results using deep learning. Here, we mainly introduce three types of basic approaches.
(i) A generative learning approach proposed to generate more data from real data distribution with
generative adversarial networks \cite{Odena2016SemiSupervisedLW,cap2020leafgan} or variational auto-encoders \cite{Kingma2014SemisupervisedLW}. 
(ii) Pseudo-Labeling based methods firstly train a deep model on labeled data in a supervised manner and use it to predict pseudo-labels from unlabeled data \cite{rizve2021defense,lu2021semi,liao2020weakly,Fan2022PrivateSF}. (iii) Furthermore, data augmentation methods such as MixUp \cite{zhang2018mixup} and MixMatch \cite{Berthelot2019MixMatchAH}, propose to blend two images and their respective labels at pixel-level, demonstrating superior performance. Instead of directly mixing pixels, CutMix \cite{Yun2019CutMixRS} and ClassMix \cite{Olsson2021ClassMixSD} mix object-label information, making them applicable to semantic segmentation.

Despite all the efforts made by prior works, it is, however, difficult to apply SSL for our drone perception task. Since ground viewpoint and drone flying viewpoints have significant differences, pseudo-labeling methods tend to generate wrong pseudo-labels. Similarly, data augmentation methods cannot generate effective training pairs of images and semantic maps. Besides, due to the lack of labeled data of flying viewpoints, generative approaches cannot produce realistic flying data.
In this paper, we propose the nearest-neighbor pseudo-labeling and MixView to deal with the viewpoint difference.


\section{Technical Approach}
\label{sec_method}

In this section, we present our progressive self-distillation framework for SSL-based drone perception. 
Our goal is to train a viewpoint invariant model that enables a drone to perceive precisely from any flying height in low altitude\footnote{In high altitudes, objects eventually tend to be extremely small and deformed in a camera coordinate, and viewpoint difference hardly further degrades the perception performance. In this case, supervised learning is considered the only effective method.}. Instead of assuming a simple case where labeled data is evenly distributed in each viewpoint, we consider a challenging scenario where only data collected from the ground viewpoint are labeled and all data captured from other viewpoints are unlabeled. 
We pose this problem as a heterogeneous multi-robot knowledge distillation problem and perform semi-supervised ground-to-aerial knowledge transfer by distilling the perception ability from a UGV to a drone.

Fig.~\ref{fig_method} demonstrates the overall information flow in our progressive SSL framework.
The labeled samples $(x_1,y_1)$ of ground viewpoint are utilized to train a network for UGV perception.
The network is denoted by $N_1$
and will be a baseline used for comparison throughout the paper.
We then predict pseudo-labels of data from the nearest neighbor viewpoint $h_2$ with $N_1$.
The resulting pseudo-labels sever as supervision for training at $h_2$.
In addition, we propose MixView that mixes data from $h_1$ and $h_2$ to generate augmented training samples. The same learning strategy is then applied to $h_3$.
After $n$ progressive applications on heights $h_1$ to $h_n$, we will obtain the final model $N_n$ that is able to handle any image taken from  $h_1$ to $h_n$.

In the rest of this section, we provide the details of each important component in our framework. 
Specifically, we discuss viewpoint sampling when applying our method in Sec.~\ref{method-1}. In Sec.~\ref{method-2}, we describe how to take advantage of the nearest neighbor viewpoint and apply pseudo-labeling to it. In  Sec.~\ref{method-3}, we further propose MixView to better overcome viewpoint differences. 
In Sec.~\ref{method-4}, we present the progressive self-distillation to obtain the final model.

\subsection{Dense Sampling of Viewpoint}
 \label{method-1}
 One essential key to achieving a superior performance of our method is the strategy of dense sampling of viewpoints. As we handle significant viewpoint differences by gradually leveraging intermediate viewpoints,
the intermediate viewpoints play a role in viewpoint assistance. Hence, the more assistant viewpoints we use, the better accuracy we will have. 

Formally, given the maximum flight height $h$, we propose to sample flight data with a uniform interval $h/n$ by
 \begin{equation} \label{eq_sampling}
   h_i = h/n \times i
\end{equation}
where $i \in [1,n]$.
Eq.~\ref{eq_sampling} divides the flight range $[0,h]$ into 
 $[h_1, h_2,\ldots, h_n]$, where $h_1 < h_2< \ldots < h_n$.


We assume only images of the ground viewpoint are labeled, while those of the other viewpoints are unlabeled. For simplicity,  $\mathcal{X}_1$ denotes a labeled set of the ground viewpoint, while $\mathcal{U}_i$ and $\mathcal{X}_i$ denote the unlabeled set and its pseudo-labeled version of the flying viewpoint $h_i$, respectively, throughout the paper.

\subsection{Nearest Neighbor Pseudo-Labeling}

 \label{method-2}
 
\begin{figure}[t]
\centering
\includegraphics[width=0.39\textwidth]{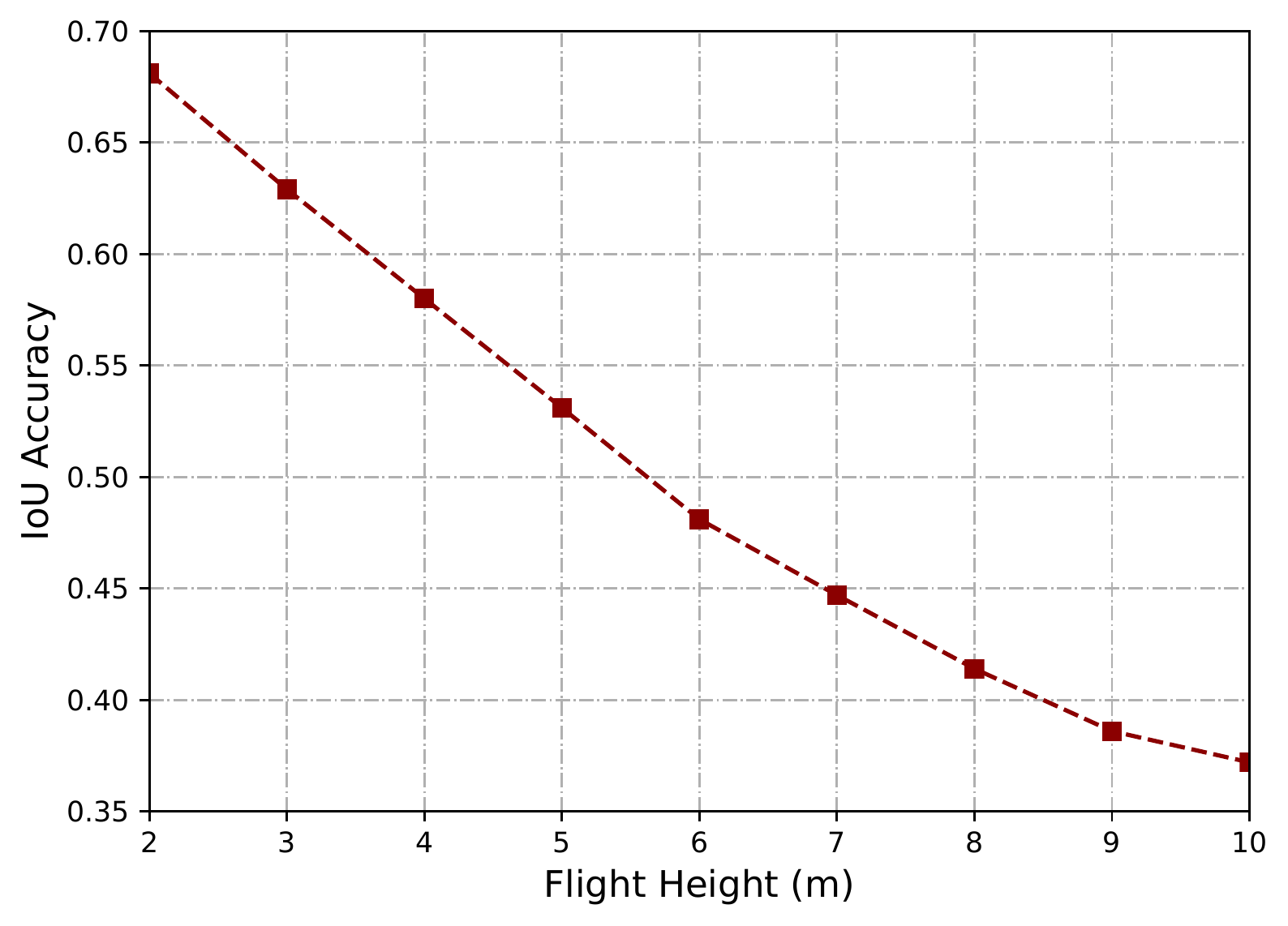}
\vspace{-2mm}
\caption{The mean IoU accuracy of a  network model trained on data collected from the ground viewpoint for different flight heights.}
\label{fig_config}
\end{figure}

 Fig.~\ref{fig_config} shows the accuracy of $N_1$ for various flight heights. It is obvious that
 the viewpoint difference deteriorates UGV's performance. On the other hand, close observation shows a small performance gap between two neighbor viewpoints, e.g., flight height of 2 meters and 3 meters. 
  It is, therefore, reasonable to exploit SSL between a viewpoint and its nearest neighbor.
  We propose {\it the nearest neighbor pseudo-labeling} that applies pseudo-labeling to the nearest viewpoint.
  
Let 
$(x_1, y_1)$  be a pair at $h_1$, and $x_2$ be an image at $h_2$, where $(x_1, y_1) \in \mathcal{X}_1$ and $x_2 \in \mathcal{U}_2$.
A straightforward way to infer a label at $h_2$ is utilizing $N_1$ learned at $h_1$ by
\begin{equation}\label{NNPL}
   \hat{y}_2 = N_1(x_2)
\end{equation}
where $\hat{y}_i$ is the one-hot label of $x_i, \forall i,$ called ``pseudo-label" as it provides an  approximation to its correct semantic map.

Eq.~(\ref{NNPL}) allows an implicit entropy minimization to improve SSL and is effective due to small viewpoint differences. 
 Then, we can learn a better model at $h_2$ by mixing data from the nearest neighbors of $h_1$ and $h_2$. 
The model $N_2$ is learned through minimizing the objective by
 \begin{equation}
\underset{N_2} \min \  \frac{1}{|\mathcal{X}_{1}|}  \! \sum_{(x_1, y_1)  \! \in \mathcal{X}_1} \! \mathcal{H}(y_1, N_2(x_1)) \! + \frac{1}{|\mathcal{U}_2|} \sum_{x_2 \in \mathcal{U}_2} \mathcal{H}(\hat{y}_2, N_2(x_2)) 
\end{equation}
where $\mathcal{H}$ is the cross entropy function.
The pseudo-labels are obtained at $h_i$ by
 \begin{equation}
   \hat{y}_i = N_{i-1}(x_i)
\end{equation}
where $N_{i-1}$ denotes a  model trained at $h_{i-1}$. 
To avoid confusion, we simplify our notation by stating that the ground truths $y_1$ at the ground viewpoint are equivalent to $\hat{y}_1$ for the rest of the paper.
Then, the optimization objective for learning $N_i$ can be written as
 \begin{equation}
 \begin{aligned}
    & \underset{N_i} \min \    \frac{1}{|\Psi_{i}|} \sum_{(x,\hat{y}) \in \Psi_i} \mathcal{H}(\hat{y}, N_i(x)) \\
    & \text{s.t.}\ \ \   \mathcal{X}_i = N_{i-1}(\mathcal{U}_i)  \\
    & \qquad 
    \Psi_i= \bigcup _{k=1}^i\mathcal{X}_k
 \end{aligned}
\end{equation}

\subsection{MixView}

\label{method-3}

Ideally, an object, e.g., a car, should be correctly recognized from different flight heights despite its appearance/shape being different among viewpoints. This can be readily implemented if we have ground truths.
In our setting, we propose MixView to ensure this viewpoint invariance. We argue that an object should also be correctly recognized if we manually move it to another viewpoint. 
With this intuition, we apply MixView to mix data samples among different viewpoints at the object level to generate augmented data.

MixView can be seen as a variant of ClassMix \cite{Olsson2021ClassMixSD} to handle SSL under viewpoint difference.
Instead of augmenting only the unlabeled set, we make an improvement of the original ClassMix where we blend the labeled set and unlabeled set to generate viewpoint robust images. 
We apply MixView when training $N_i$ in an online fashion, and the model gradually learns to recognize objects under viewpoint difference. We formulate the MixView by

\begin{equation}
\begin{gathered}
\vspace{1mm}
    (x'_i,y'_i) = MixView\left( (x,\hat{y}), (x_i,\tilde{y}_i))\right) \\
    x'_i = m \odot x_i + (1- m) \odot x \\
    y'_i = m \odot \tilde{y}_i + (1- m) \odot \hat{y}
\end{gathered}
\label{equ_mixclass}
\end{equation}
where  $(x, \hat{y}) \in \Psi_i$, $\tilde{y}_i = N_i(x_i)$, $m$ is a binary mask that randomly exchanges half of the classes between $x_i$ and $x$ to generate a new image $x'_i$. $\odot$ denotes element-wise multiplication.

The  “manifold" assumption of SSL assumes that the data space is composed of multiple lower-dimensional manifolds, and data points lying on the same manifold should have the same label \cite{vanEngelen2019ASO}.
Viewpoint discrepancy causes misclassification in the data space. To solve this problem, MixView imposes a regularization to viewpoint differences and helps to move misclassified data points to the correct manifold.

\subsection{Progressive Self-Distillation}
 \label{method-4}
 
We formally describe our progressive learning strategy that integrates the nearest neighbor pseudo-labeling and MixView. 
Specifically, given a labeled set $\mathcal{X}_1$ and trained model $N_1$ at $h_1$, as well as a set of unlabeled samples $\mathcal{U}_2, \mathcal{U}_3, ..., \mathcal{U}_n$ taken from $h_2, h_3, ..., h_n$, respectively, we will progressively transfer the perception ability from $N_1$ to various flying heights.

We train the model $N_i$ with the labeled set  ${\Psi_{i}= \bigcup _{k=1}^i\mathcal{X}_k}$ 
and the augmented data of $\mathcal{U}_i$. 
 We train the final model $N_n$ looping over all viewpoints by
\begin{equation} \label{final_loss}
\begin{aligned}
& \underset{N_n} \min \ \mathcal{L} \\  
& \text{s.t.}\ \ \   \mathcal{X}_i = N_{i-1}(\mathcal{U}_i) \\
& \qquad  \Psi_i= \bigcup _{k=1}^i\mathcal{X}_k \\
& \qquad  (x'_i,y'_i) = MixView\left( (x,\hat{y}), (x_i,\tilde{y}_i))\right),  
\end{aligned}
\end{equation}
where 
\begin{equation} \label{final_loss}
\mathcal{L} \! = \! \sum_{i=2}^n \! \Big( \! \frac{1}{|\Psi_i|}  \! \sum_{(x,\hat{y})  \! \in \Psi_i}  \! \mathcal{H}(\hat{y}, N_i(x))  \!  + \lambda \ \frac{1}{|\mathcal{U}_i|} \! \sum _{x' \! \in \mathcal{U}_i} \mathcal{H}(y', N_i(x')) \Big) 
\end{equation}
and $\lambda \in [0,1]$ is a hyperparameter varying from 0 to 1 during the optimization.

\begin{algorithm}[t]  
  \caption{Algorithm of our progressive SSL framework.}  
  \label{alg_leaning_G}  
  \begin{algorithmic}[1]  
    \Require  
      $\mathcal{X}_1$: A labeled set of image pairs $(x_1,y_1)$ at the ground viewpoint $h_1$;
      $\mathcal{U}_i$: An unlabeled set from the viewpoint $h_i$, $\forall i=2,3,\ldots,n$;  
      $N_1$: The baseline model for ground viewpoint perception.
     \renewcommand{\algorithmicrequire}{\textbf{Hyperparameters:}}
    \Require 
    Initial learning rate: $0.01$,  weight decay: $1e^{-4}$, number of training steps: \it{iterations}.
    \Ensure  
      $N_n$: The final model trained on images captured from viewpoints $h_1$ to $h_n$.  
    \State Freeze $N_1$;
    \Statex \LeftComment{0} {\color{gray} \% Progressive Learning \%}
   \For{$i=2$ to $n$}           
   
   \Statex \LeftComment{1} {\color{gray} \% Applying the nearest neighbor pseudo-labeling \%}
    \State   $\mathcal{X}_i = N_{i-1}(\mathcal{U}_i)$  
        \State $\Psi_i= \bigcup _{k=1}^i\mathcal{X}_k $
        \State Initialize $N_i$;
        \For{$j$ = 1 to $iterations$}             
            \State Set gradients of $N_i$ to 0;
            \State Select $(x,\hat{y})$ from  $\Psi_i$ and  $x_i$ from $\mathcal{U}_i$;
            \State  $\tilde{y}_i = N_i(x_i)$; 
            \Statex \LeftComment{2} {\color{gray} \% Applying MixView \%}
            \State $(x'_i,y'_i)$ = MixView$\left( (x,\hat{y}), (x_i,\tilde{y}_i))\right)$;  
            \State Calculate the loss $\mathcal{L}$ with Eq.~(\ref{final_loss});
             \State Backpropagate $\mathcal{L}$;
             \State Update $N_i$; 
        \EndFor
        
    \Statex   \LeftComment{1} {\color{gray} \% Updating $\mathcal{X}_i$ \%}
        \State      $\mathcal{X}_i = N_i(\mathcal{U}_i)$ 
         \EndFor           
  \end{algorithmic}  
\end{algorithm}

\begin{figure*}[t]
\centering
\includegraphics[width=0.98\textwidth]{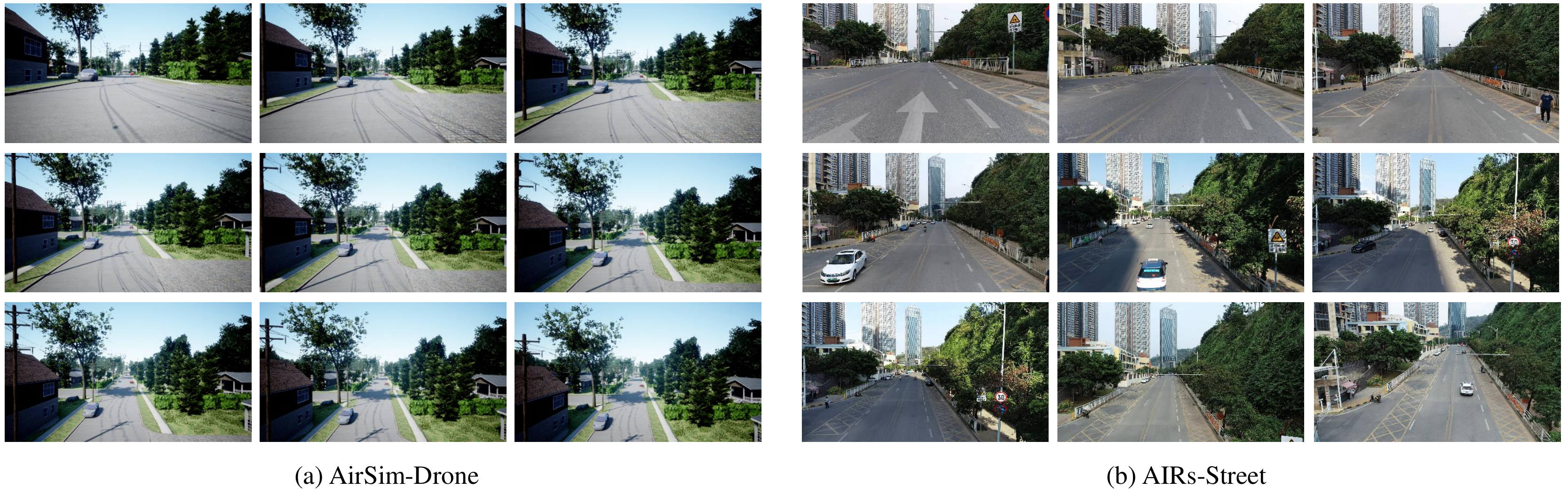}
\vspace{-2mm}
\caption{Examples of selected images from our datasets. (a) shows nine images of the flight heights from 1 meter to 9 meters in AirSim-Drone, and similarly, (b) shows nine images of the flight heights from 1 meter to 9 meters in AIRs-Street.}
\label{fig_selected_imgs}
\end{figure*}

\section{Datasets for Drone Perception}
\label{sec_dataset}
In this section, we introduce our datasets for the evaluation of the proposed method. 
As there is no prior dataset that can be applied to our problem, to better verify the performance of the proposed method and inspire more future explorations, we create both a synthesized AirSim-Drone and a real-world AIRs-Street, respectively. The former is collected from the Microsoft AirSim and is used for evaluating our method in an ideal case, and the latter is captured in a city street scenario and is used to quantify our method in the real world. We will present the details of each dataset for the rest of this section.

\subsection{AirSim-Drone}
We synthesize the AirSim-Drone from the AirSim Neighborhood environment where we capture the data at the ground viewpoint with a car, and flight data with a drone. The flight route of the drone follows a predefined map
while objects and lighting conditions such as illumination and weather, are all fixed.
We collect the flight data while varying flight height to  2, 3, 4, 5, 6, 7, 8, 9, and 10 meters, respectively. 
Since we can stably and accurately control the drone in this simulated environment, the viewpoint discrepancy is only derived from the flying height.
In addition, we collect a test sequence from a different area to the above sequences. We fly the drone at random flight heights. The test sequence is called ``uav\_random" and is used to evaluate the generalization performance.

There are 9 semantic attributes in total, including ``Plant", ``Building", ``Road", ``Sky", ``Car", ``Ground", ``Fence", ``Pole", and ``Others". Each image has a resolution of $768 \times 432$.
Detailed information of the dataset, including the flying height, number of images, and labeled semantic maps are given in Table~\ref{aisim_drone}. Note that the start and end time of video recording have to be manually determined, and thus the number of captured images in each sequence is slightly different. Since semantic maps can be readily obtained from the simulator, we have ground truth semantic annotations for all RGB images.
Only ground truths of the ground viewpoint are used for training, and other viewpoints are used for validation.

\subsection{AIRs-Street}
AIRs-Street is a real-world dataset in which images are captured in a street scenario.
To precisely measure the distance from the ground, we use an advanced DJI Phantom Pro 4 which is equipped with a 3D TOF for data acquisition. The drone has to be manually controlled to maintain its flying height following a predefined route. Thus, there are some unavoidable deviations in terms of flying height. We carefully fly the drone with a fairly slow forward speed, such that the deviation of the flight height is guaranteed within $\pm 0.5$ meters.
Similar to the AirSim-Drone, we take images of different flight heights from 1 meter to 9 meters with an interval of 1 meter, and we have 9 sequences in total. 
The resolution of the captured images is $ 1920 \times 1080$.

Due to the difference in flying speed, the number of captured images is also different among viewpoints. Despite that, all sequences have the same start point and endpoint.
For the sequence of the ground viewpoint, we uniformly select 117 frames to annotate and use for training. 
 For each of the flight sequences, we uniformly select 25 frames over the video length to annotate and use them for evaluation \footnote{Owing to lack of additional labeled images, we directly use these labeled flight data for evaluation.}.
 We manually annotate the pixel-wise attribute labels. In our dataset, there are 13 semantic classes, including  ``Plant", ``Building", ``Road", ``Sky",  ``Car",  ``Sidewalk", ``Pedestrian", ``Motorcycle",  ``Wall", ``Fence", ``Traffic Sign", ``Traffic Light", and ``Others".
More detailed information is given in Table~\ref{airs_drone}.

\begin{table}[t]
\begin{center}
\caption{The detailed information of our AirSim-Drone dataset.}
\label{aisim_drone}
\begin{tabular}
{p{0.06\textwidth}<{\centering}m{0.1\textwidth}<{\centering}m{0.1\textwidth}<{\centering}m{0.11\textwidth}<{\centering}}
\hline
Sequences & Flight height & All samples & Labeled samples\\   \hline
car01 & 1 & 1947 & 1947\\
uav02 & 2 & 1057 & 1057\\
uav03 & 3 & 1057 & 1057\\
uav04 & 4 & 1043 & 1043\\
uav05 & 5 & 1046  & 1046\\
uav06 & 6 & 1047  & 1047\\
uav07 & 7 & 1048 & 1048\\
uav08 & 8 & 1048 & 1048\\
uav09 & 9 & 1037 & 1037\\
uav10 & 10 & 1037  & 1037\\
\hline
uav\_random & 2 to 10 &1143 & 1143 \\
\hline
\end{tabular}
\end{center}
\end{table}

\begin{table}[t]
\begin{center}
\caption{The detailed information  of our AIRs-Street dataset.}
\label{airs_drone}
\begin{tabular}
{m{0.06\textwidth}<{\centering}m{0.1\textwidth}<{\centering}m{0.1\textwidth}<{\centering}m{0.11\textwidth}<{\centering}}
 \hline
Sequences & Flight height & All samples & Labeled samples\\   \hline
car01 & 1 & 770 & 117\\
uav02 & 2 & 750 & 25\\
uav03 & 3 & 717 & 25\\
uav04 & 4 & 548 & 25\\
uav05 & 5 & 591  & 25\\
uav06 & 6 & 475  & 25\\
uav07 & 7 & 428 & 25\\
uav08 & 8 & 657 & 25\\
uav09 & 9 & 623 & 25\\
\hline
\end{tabular}
\end{center}
\end{table}

\begin{table*}[t]
\begin{center}
\caption{The mean IoU accuracy of our approach and other methods on the AirSim-Drone dataset.}
\label{res_airsim_iou}
\begin{tabular}
{m{0.15\textwidth}<{\raggedleft}|m{0.05\textwidth}<{\centering}m{0.05\textwidth}<{\centering}m{0.05\textwidth}<{\centering}m{0.05\textwidth}<{\centering}m{0.05\textwidth}<{\centering}m{0.05\textwidth}<{\centering}m{0.05\textwidth}<{\centering}m{0.05\textwidth}<{\centering}m{0.05\textwidth}<{\centering}|m{0.03\textwidth}<{\centering}m{0.03\textwidth}<{\centering}}
\hline
Methods & uav02 & uav03 & uav04 & uav05 & uav06 & uav07 & uav08 & uav09 & uav10 & mean, & \hspace{1mm} std\\  \hline
 Ground-only &0.672 &0.616 &0.561 &0.519 &0.479 &0.446 &0.418 &0.388 &0.365 &0.496 &0.105\\
Pseudo-Labeling \cite{lee2013pseudo}   &0.639  &0.578 &0.530  &0.485  &0.444 &0.404 &0.369 &0.337 &0.323  &0.457 &\textbf{0.011}\\
ClassMix \cite{Olsson2021ClassMixSD}    &0.209 &0.197  &0.190  &0.183 &0.180  &0.178  &0.176 &0.174 &0.173 &0.184 &\textbf{0.011}\\
Ours &\textbf{0.680} &\textbf{0.650} &\textbf{0.625} &\textbf{0.609} &\textbf{0.594} &\textbf{0.578} &\textbf{0.564} &\textbf{0.551} &\textbf{0.539} &\textbf{0.599} &0.047 \\
\hline
\end{tabular}
\end{center}
\end{table*}

\section{Experimental Evaluation}
\label{sec_result}

\begin{figure*}[t]
\centering  
\begin{tabular}
{m{0.05\textwidth}<{\centering}m{0.13\textwidth}<{\centering}m{0.13\textwidth}<{\centering}m{0.13\textwidth}<{\centering}m{0.13\textwidth}<{\centering}m{0.13\textwidth}<{\centering}<{\centering}m{0.13\textwidth}<{\centering}} 

10m
&\IncG[width=0.1428\textwidth]{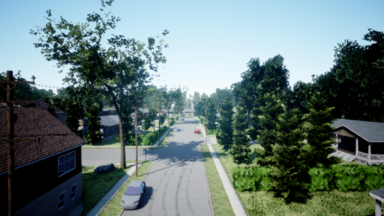}
&\IncG[width=0.1428\textwidth]{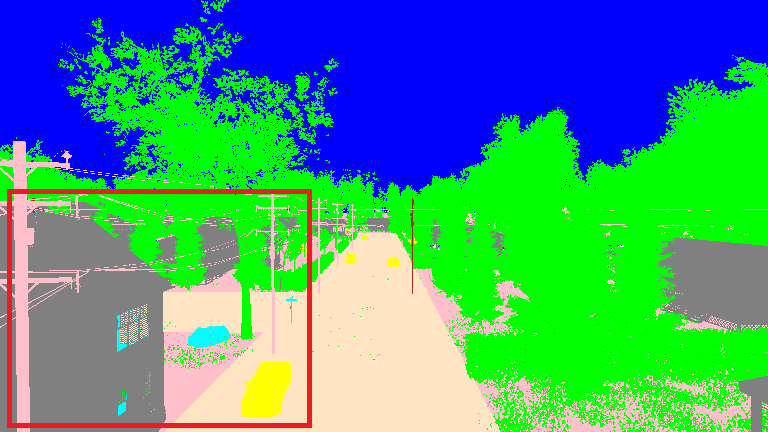}
&\IncG[width=0.1428\textwidth]{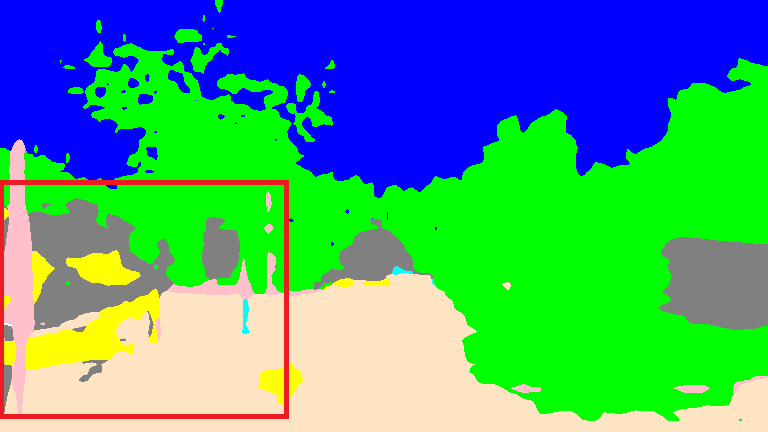}
&\IncG[width=0.1428\textwidth]{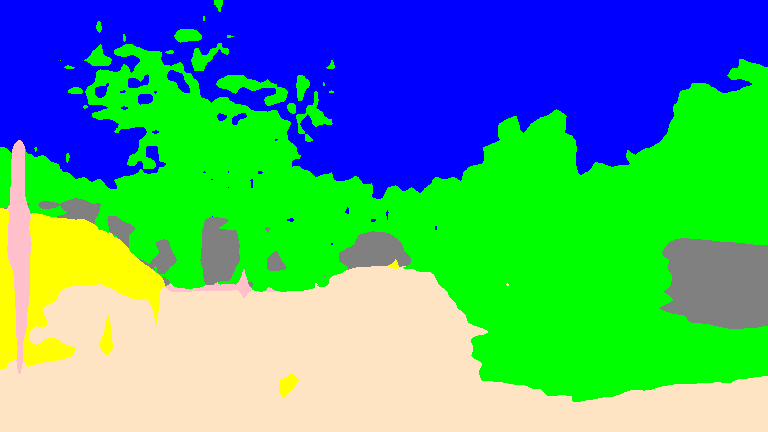}
&\IncG[width=0.1428\textwidth]{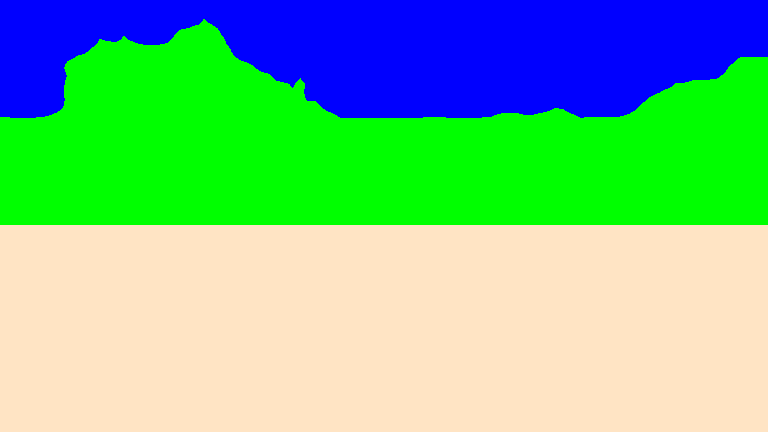}
&\IncG[width=0.1428\textwidth]{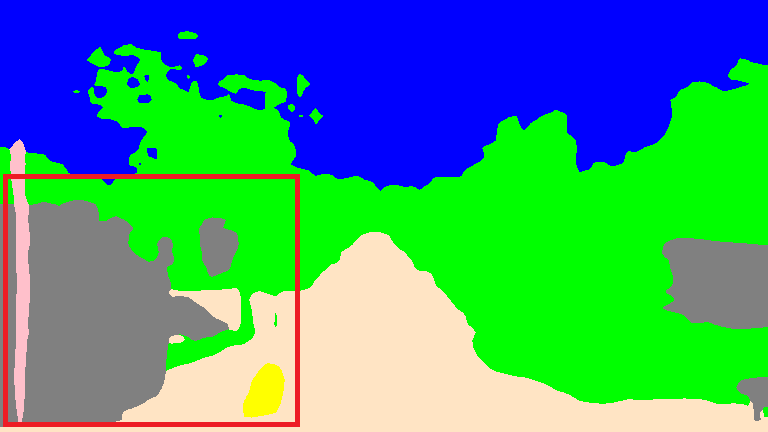}
\\
9m
&\IncG[width=0.1428\textwidth]{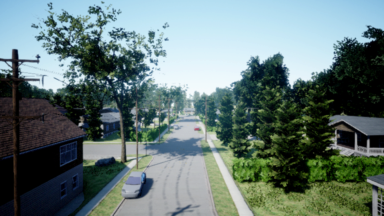}
&\IncG[width=0.1428\textwidth]{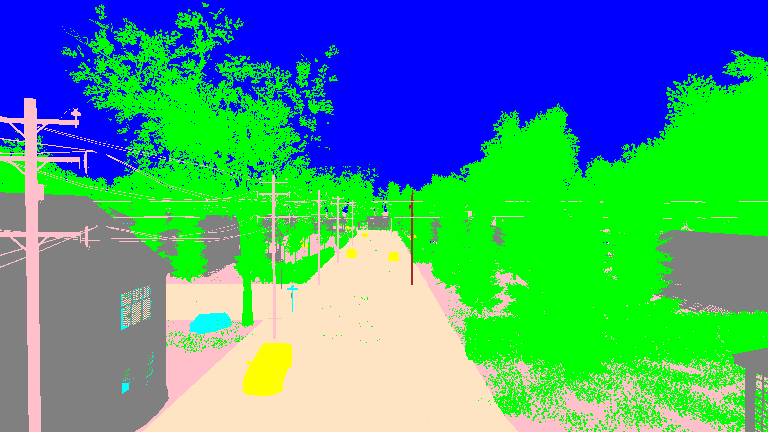}
&\IncG[width=0.1428\textwidth]{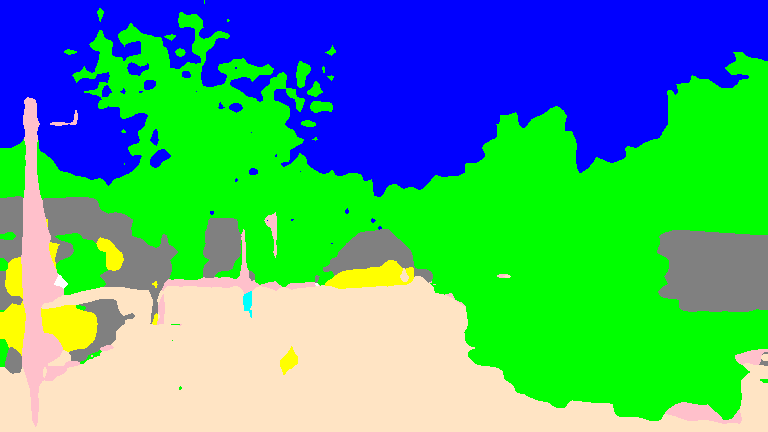}
&\IncG[width=0.1428\textwidth]{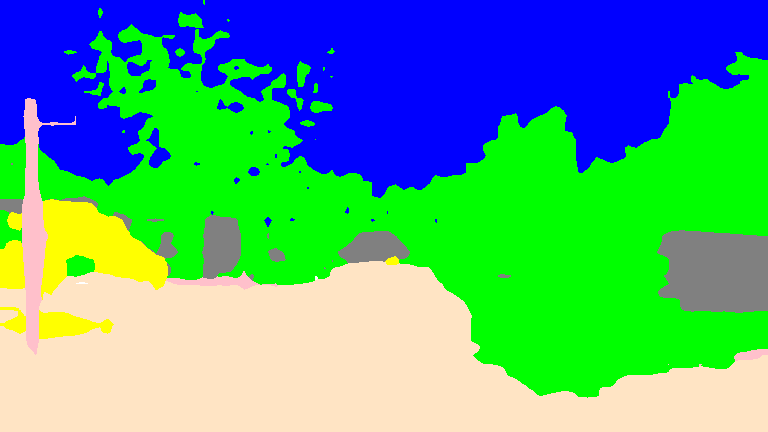}
&\IncG[width=0.1428\textwidth]{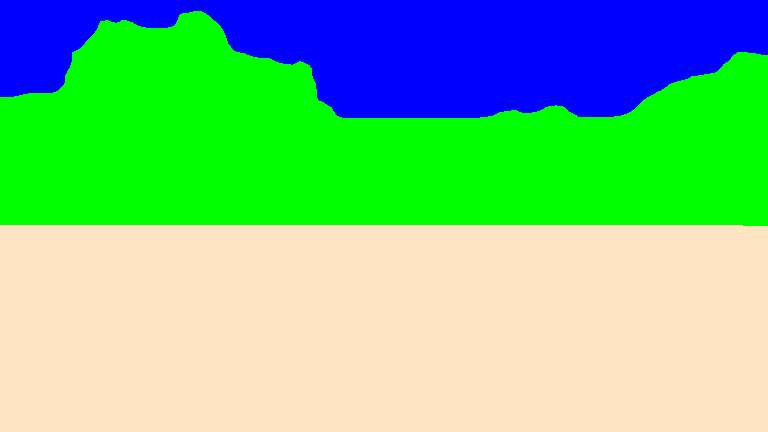}
&\IncG[width=0.1428\textwidth]{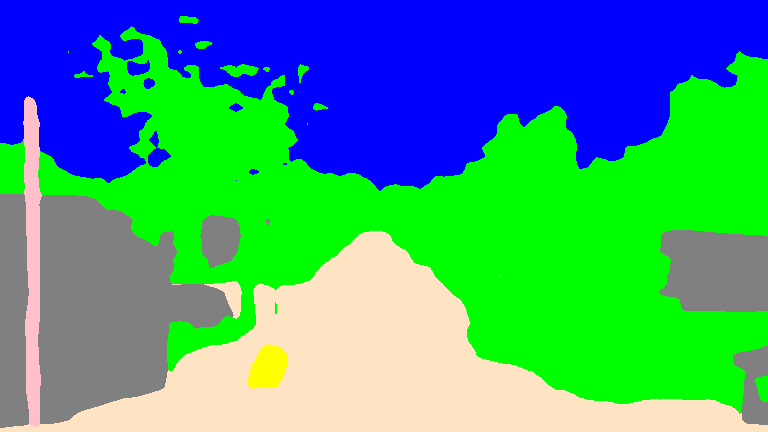}
\\
8m
&\IncG[width=0.1428\textwidth]{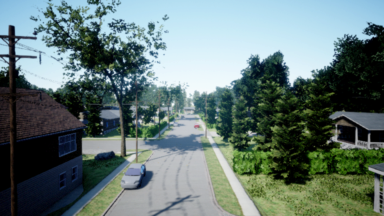}
&\IncG[width=0.1428\textwidth]{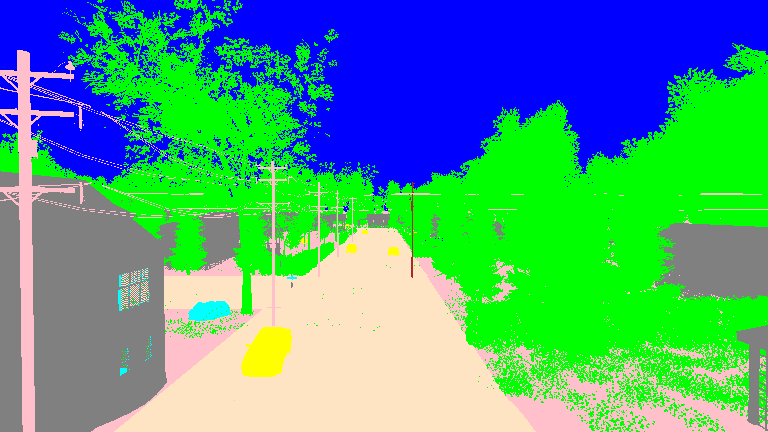}
&\IncG[width=0.1428\textwidth]{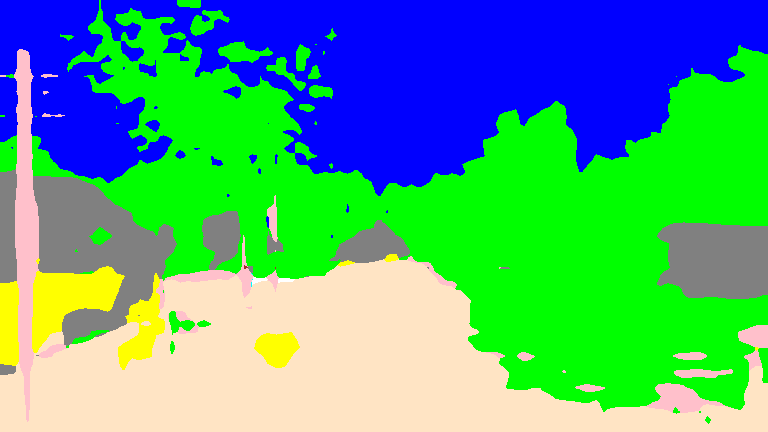}
&\IncG[width=0.1428\textwidth]{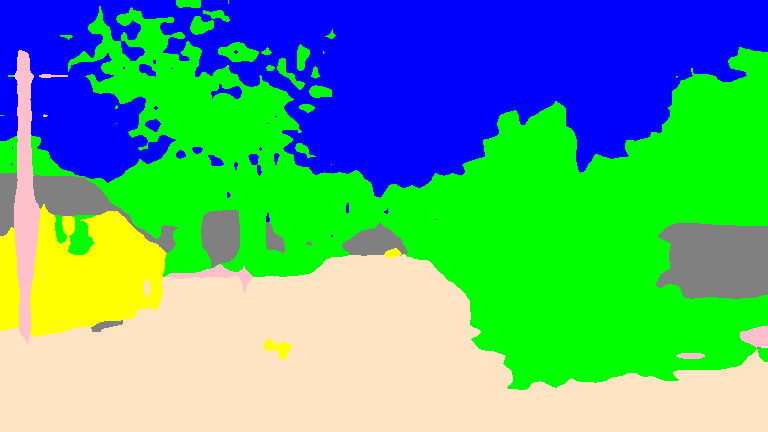}
&\IncG[width=0.1428\textwidth]{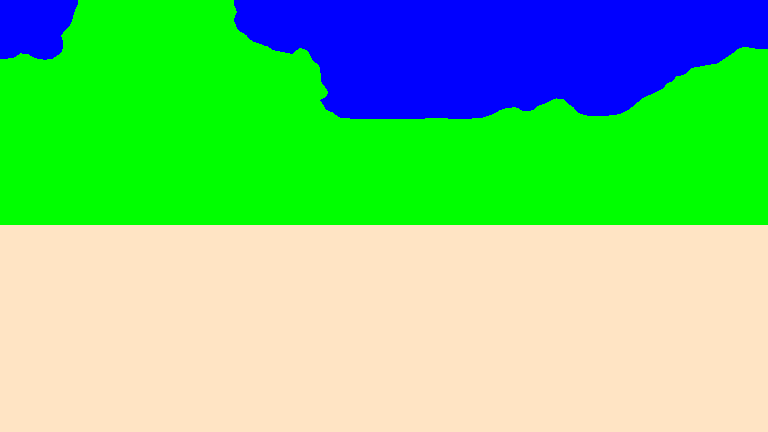}
&\IncG[width=0.1428\textwidth]{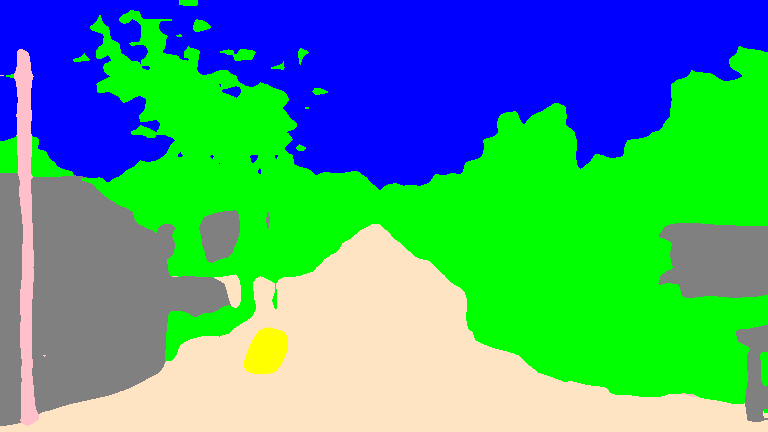}
\\

7m
&\IncG[width=0.1428\textwidth]{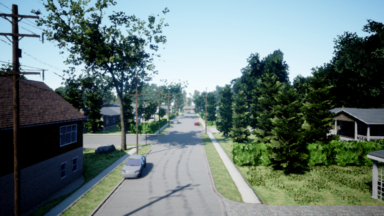}
&\IncG[width=0.1428\textwidth]{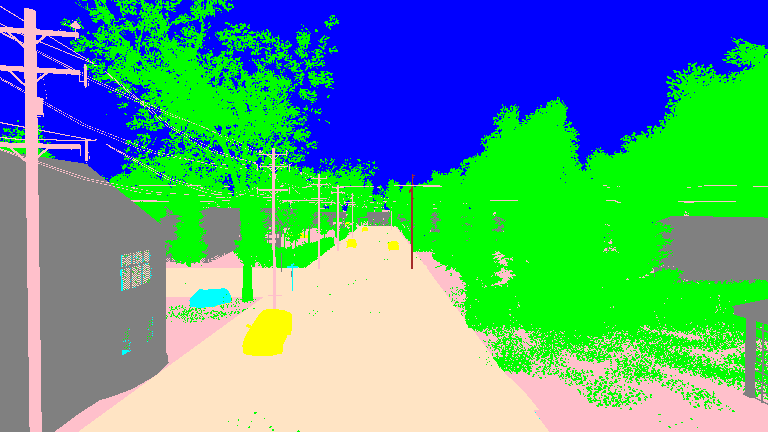}
&\IncG[width=0.1428\textwidth]{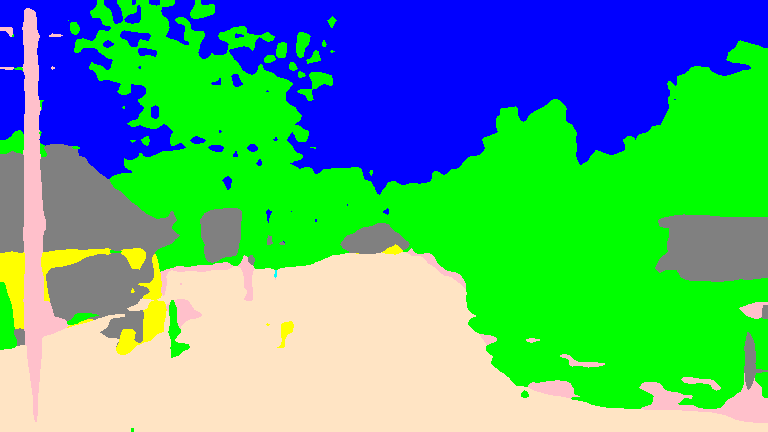}
&\IncG[width=0.1428\textwidth]{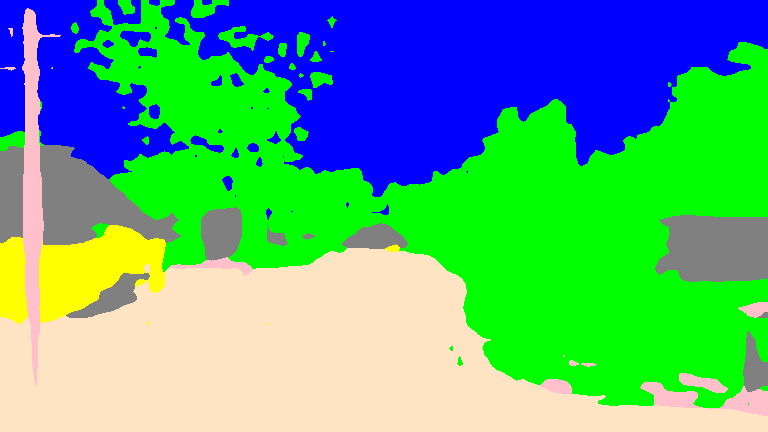}
&\IncG[width=0.1428\textwidth]{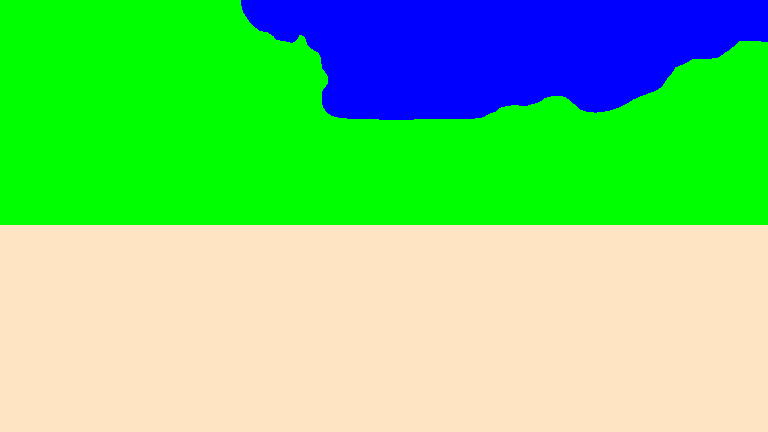}
&\IncG[width=0.1428\textwidth]{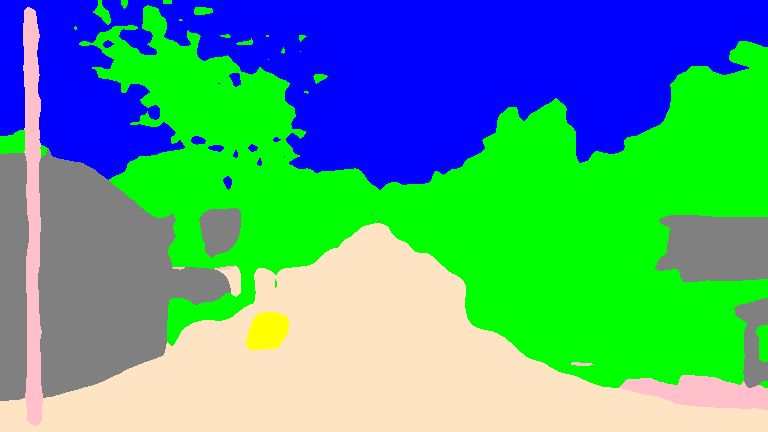}
\\

6m
&\IncG[width=0.1428\textwidth]{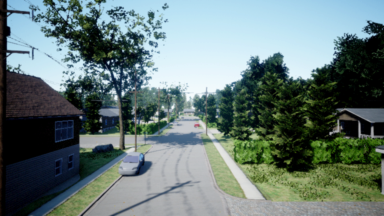}
&\IncG[width=0.1428\textwidth]{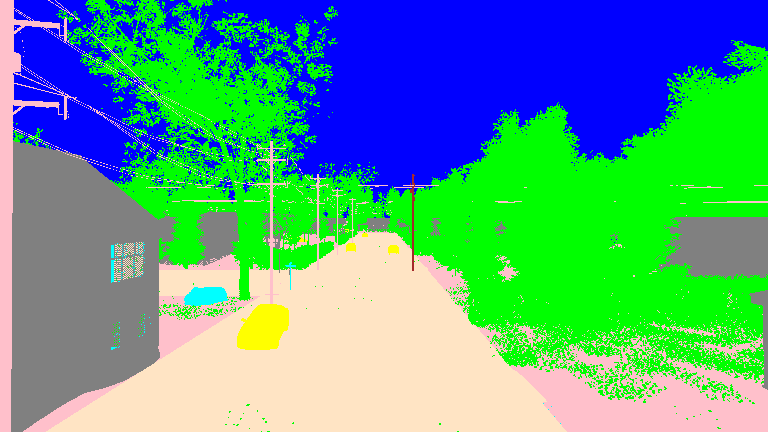}
&\IncG[width=0.1428\textwidth]{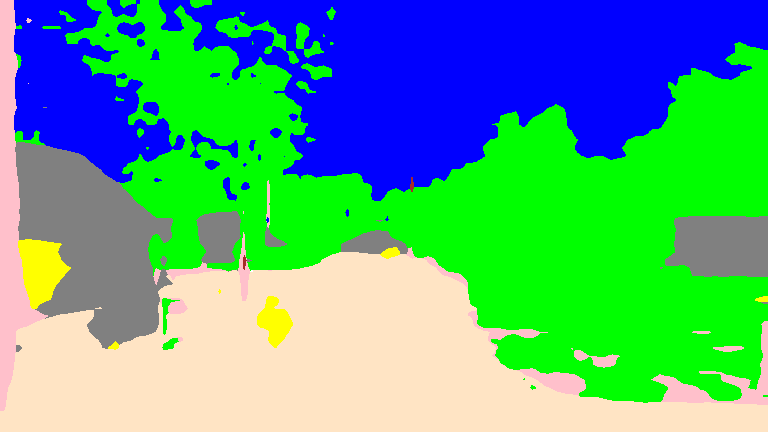}
&\IncG[width=0.1428\textwidth]{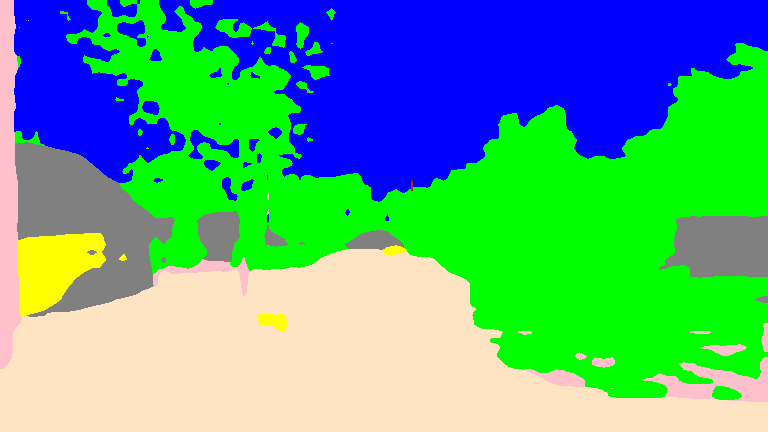}
&\IncG[width=0.1428\textwidth]{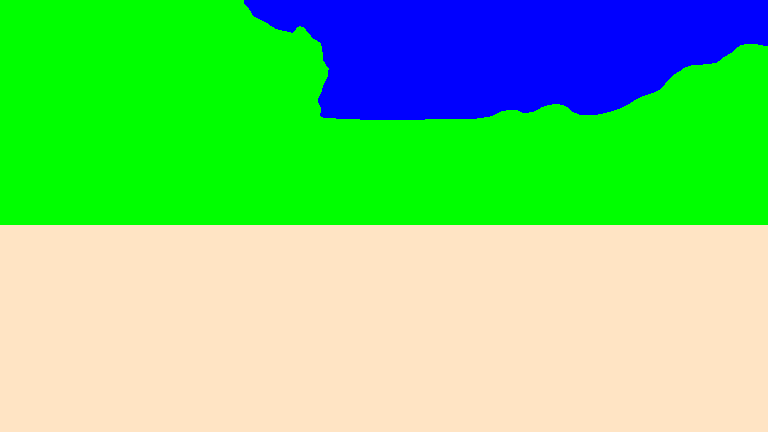}
&\IncG[width=0.1428\textwidth]{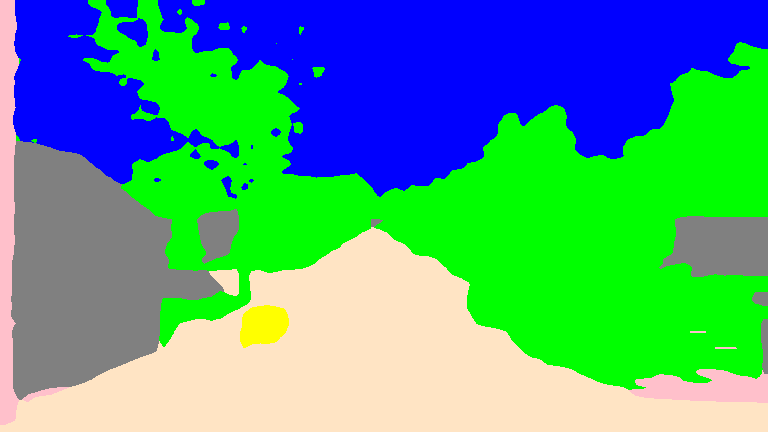}
\\

5m
&\IncG[width=0.1428\textwidth]{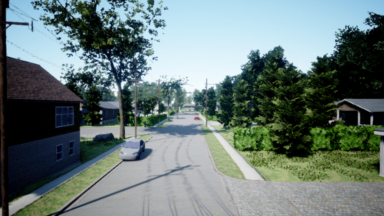}
&\IncG[width=0.1428\textwidth]{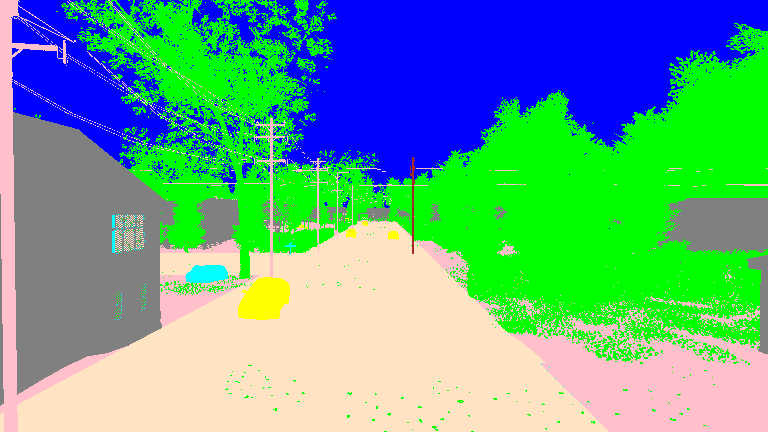}
&\IncG[width=0.1428\textwidth]{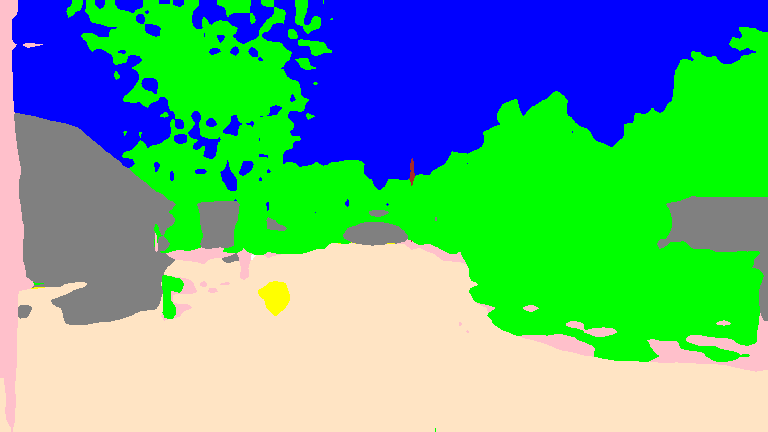}
&\IncG[width=0.1428\textwidth]{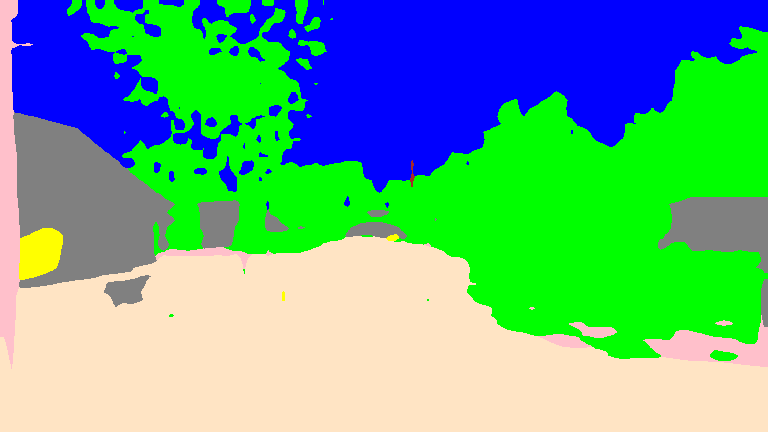}
&\IncG[width=0.1428\textwidth]{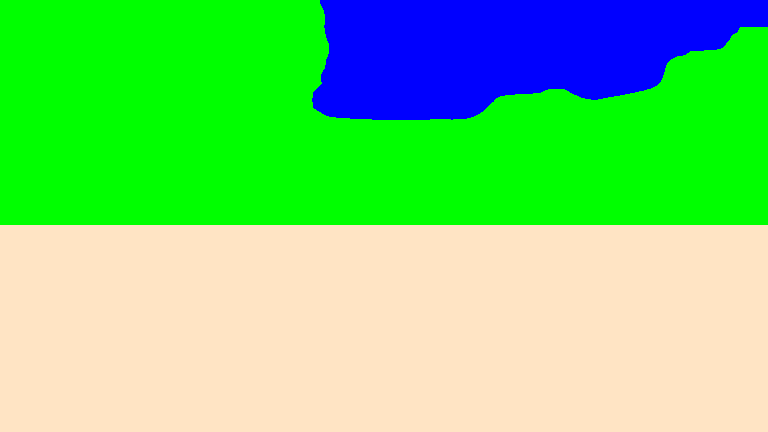}
&\IncG[width=0.1428\textwidth]{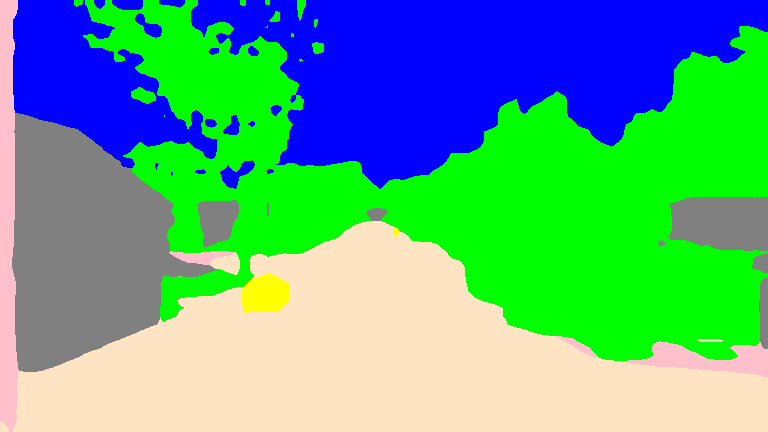}
\\
4m
&\IncG[width=0.1428\textwidth]{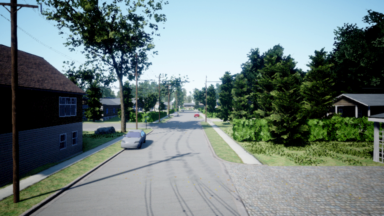}
&\IncG[width=0.1428\textwidth]{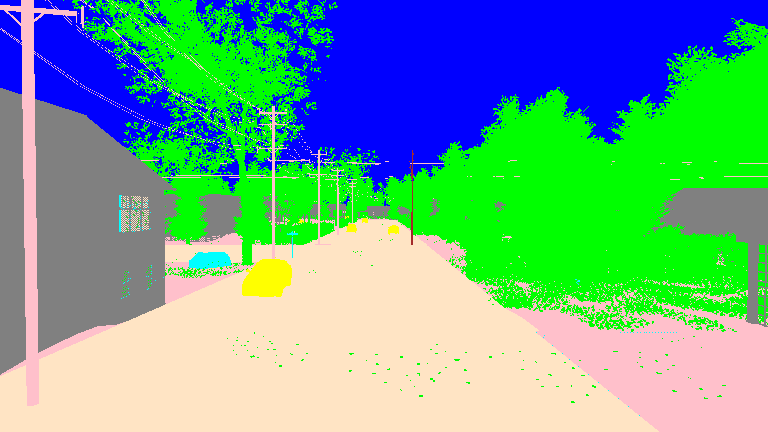}
&\IncG[width=0.1428\textwidth]{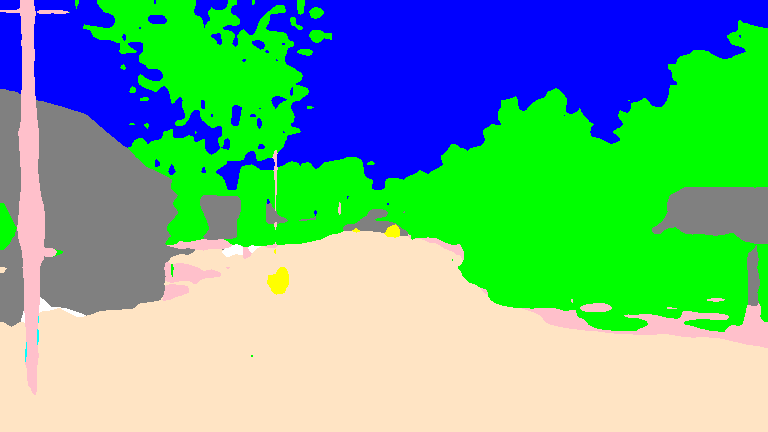}
&\IncG[width=0.1428\textwidth]{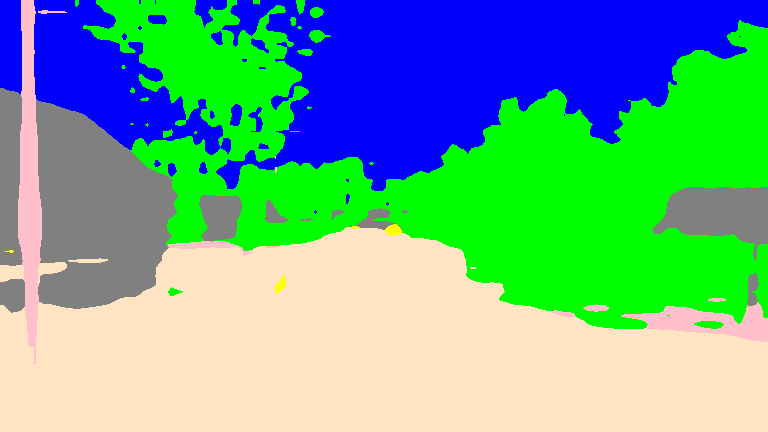}
&\IncG[width=0.1428\textwidth]{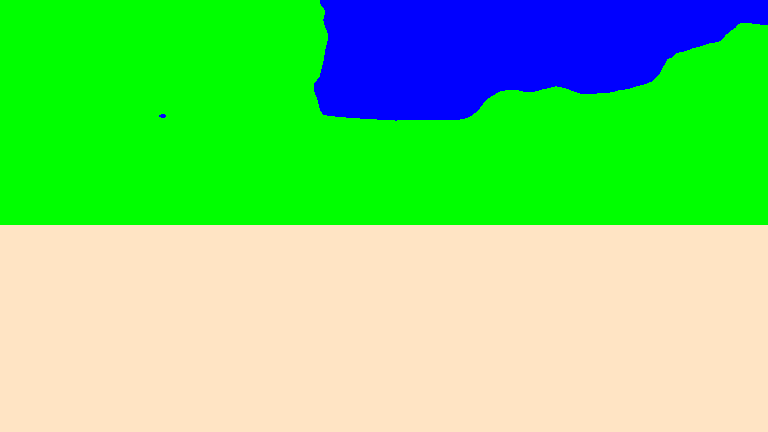}
&\IncG[width=0.1428\textwidth]{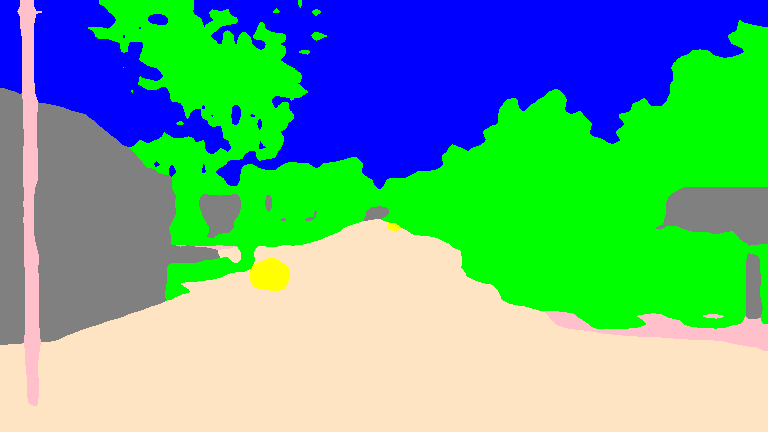}
\\
3m
&\IncG[width=0.1428\textwidth]{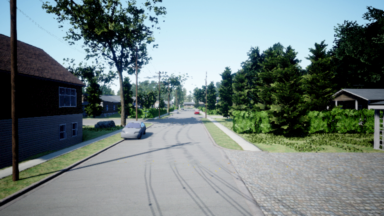}
&\IncG[width=0.1428\textwidth]{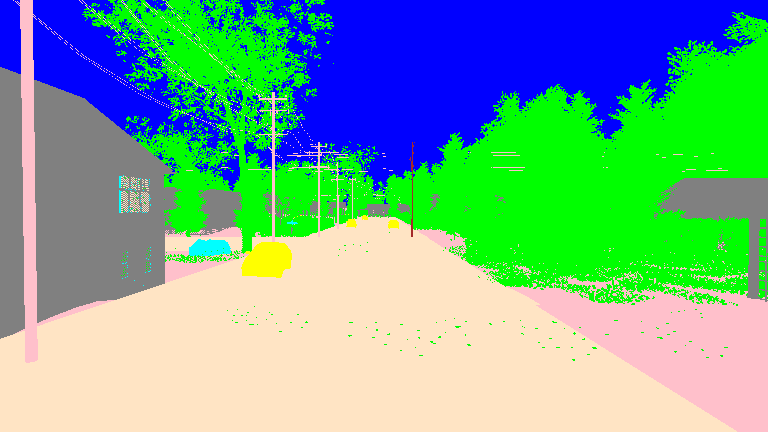}
&\IncG[width=0.1428\textwidth]{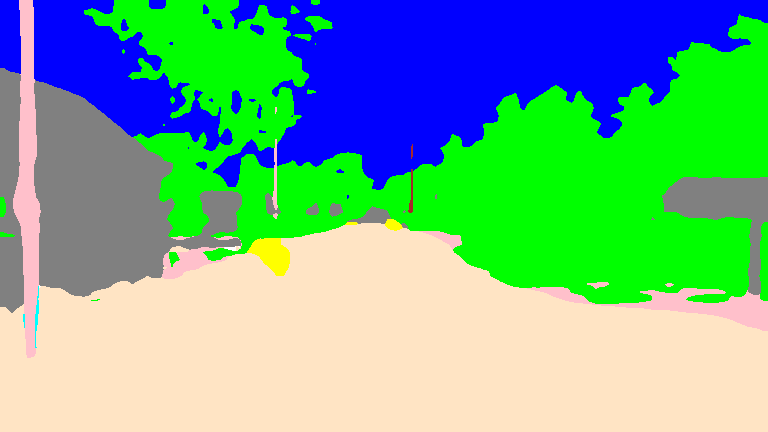}
&\IncG[width=0.1428\textwidth]{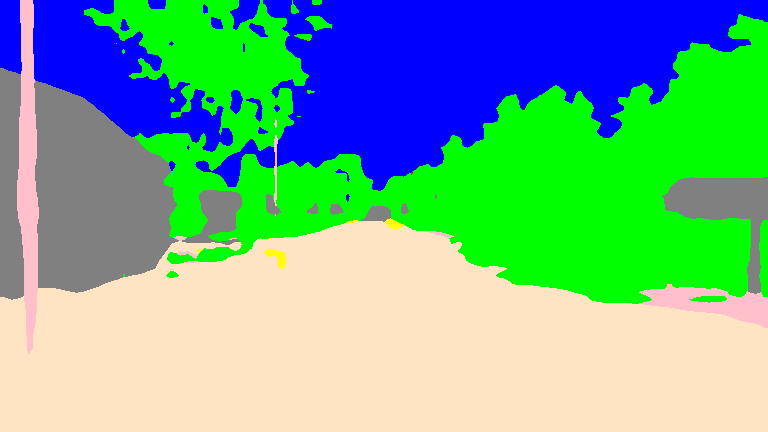}
&\IncG[width=0.1428\textwidth]{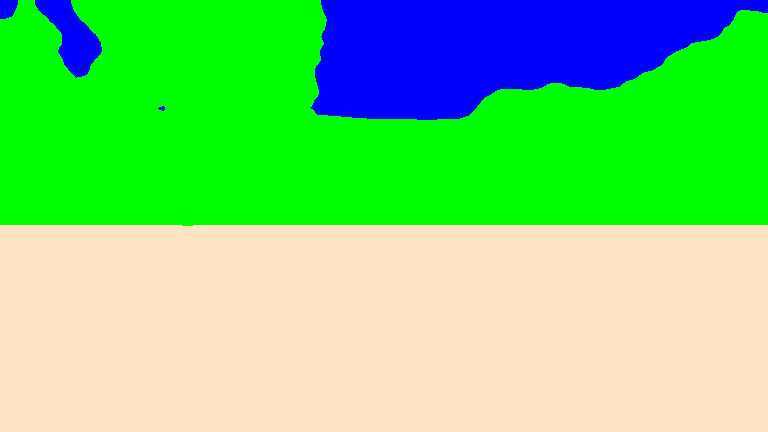}
&\IncG[width=0.1428\textwidth]{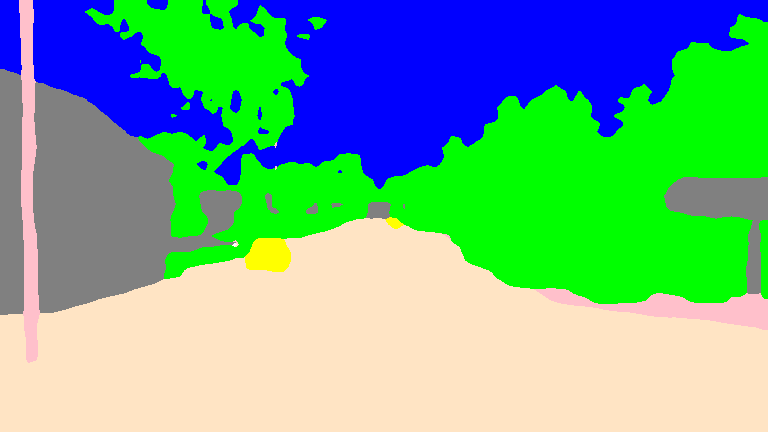}
\\
2m
&\IncG[width=0.1428\textwidth]{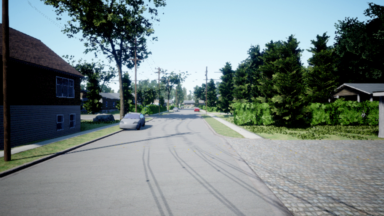}
&\IncG[width=0.1428\textwidth]{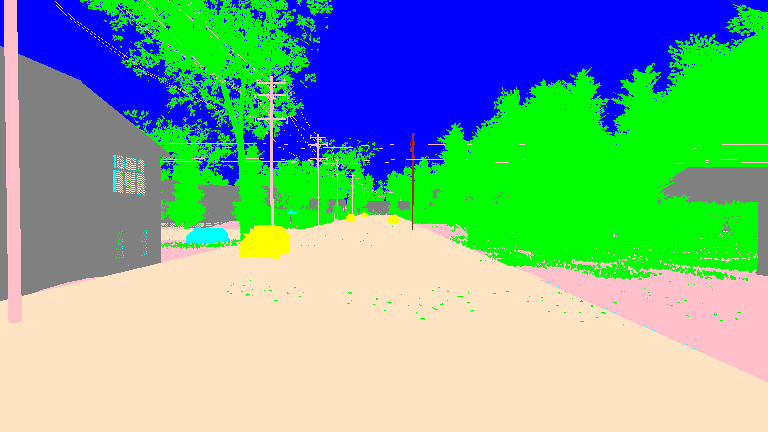}
&\IncG[width=0.1428\textwidth]{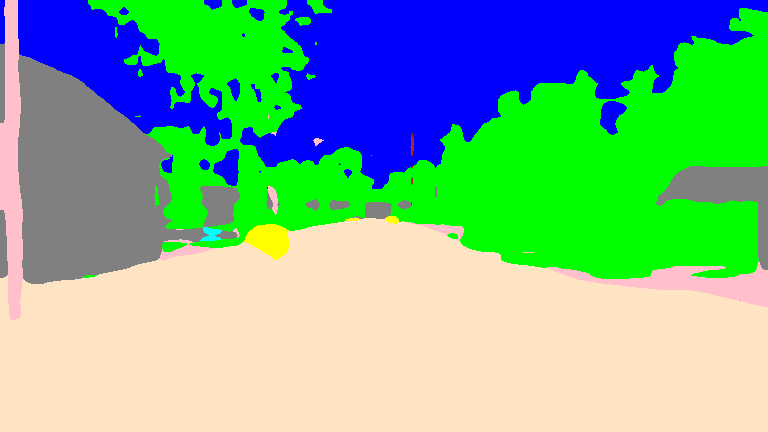}
&\IncG[width=0.1428\textwidth]{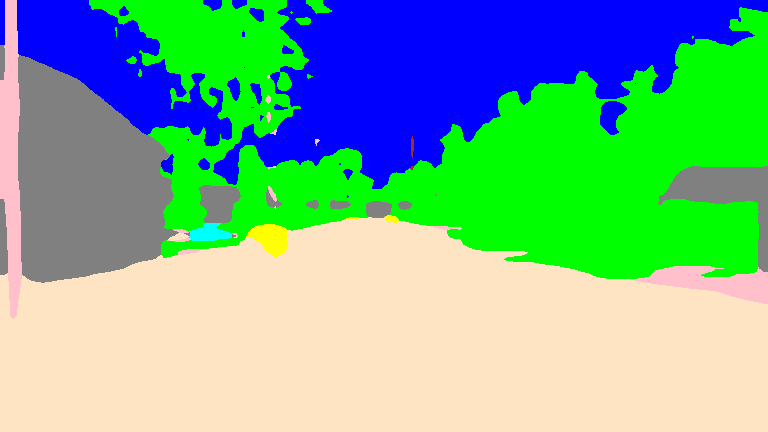}
&\IncG[width=0.1428\textwidth]{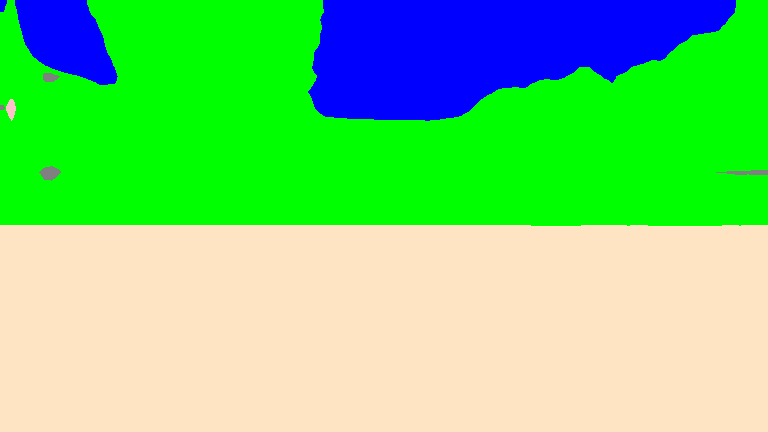}
&\IncG[width=0.1428\textwidth]{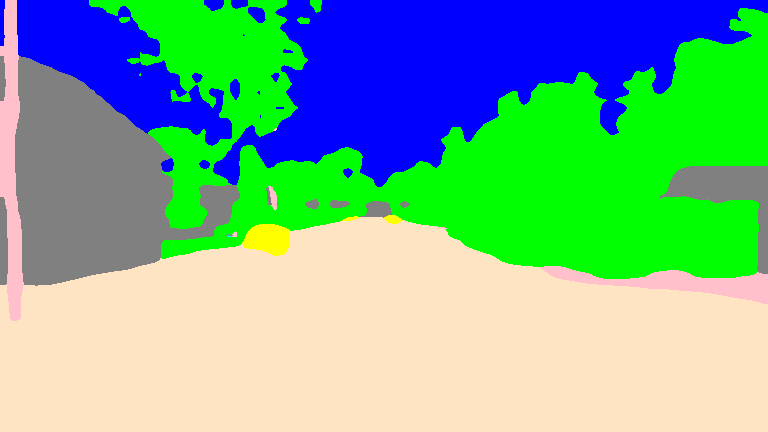}
\\
&\footnotesize{(a) Images}
& \footnotesize{(b) Ground Truths}
&\footnotesize{(c) Ground-only}
&{\fontsize{6.6}{7.92}\selectfont{(d) Pseudo-Labeling \cite{lee2013pseudo} }}
&\footnotesize{(e) ClassMix \cite{Olsson2021ClassMixSD}}
&\footnotesize{(f) Ours}
\\
\end{tabular}
\caption{Qualitative comparisons between our method and other approaches on the AirSim-Drone dataset. We show the results for different flight heights from 2 meters to 10 meters. We draw red boxes on the Ground truth and results of Ground-only and Ours at 10m for better visualization.}
\label{fig_arisim}
\end{figure*}

\subsection{Implementation Details}
\paragraph{Implementation of Our Method}
Due to the limitation of GPU memory, we change the original image resolution from $768 \times 432$ to $384 \times 216$ for AirSim-Drone, and from $1920 \times 1080$ to  $480 \times 270$ for AIRs-Street, respectively.
For the network of semantic segmentation, we use the pretrained DeepLabV3+ \cite{chen2018encoder} on the CityScape dataset \cite{Cordts2016Cityscapes}.  We modify the output layers to fit the semantic categories of our datasets.
We train $N$ for 500 epochs for each viewpoint according to Algorithm 1 with an NVIDIA GeForce RTX 2080 Ti. 
We use the SGD optimizer \cite{ruder2016overview} with a learning rate of 0.01 and apply polynomial learning rate decay during training.
We conduct all the experiments using PyTorch \cite{NEURIPS2019_9015}.

\paragraph{Baseline Methods}
Since the perception capability of the drone is distilled from the UGV, we consider the trained model at the ground viewpoint as a baseline and name it as {\it Ground-only}. 
Moreover, we additionally evaluate two previous methods of SSL on the under-explored problem for more fair comparisons. The first is Pseudo-Labeling \cite{lee2013pseudo} which is a classical approach for SSL, and the second is ClassMix \cite{Olsson2021ClassMixSD}, the current state-of-the-art for SSL-based semantic segmentation. 
Entropy minimization is employed to identify pseudo-labels from unlabeled data. Using these classical two methods as representatives, we will show through experiments that existing semi-supervised learning methods cannot cope with large viewpoint differences.

\subsection{Evaluation Metrics}
For a fair comparison of semantic segmentation, we use the metrics that were frequently employed in previous studies. We calculate the mean intersection-over-union (IoU) value. 
The mean IoU metric, also referred to as the Jaccard index, is essentially a method to quantify the percentage of overlap between the target mask and the prediction output. It is calculated by
\begin{equation}
    IoU(y, \hat{y}) = \sum_{k=1}^c \frac{ {|M(\hat{y}_k) \cap M(y_k)|}}{ {|M(\hat{y}_k) \cup M(y_k)|}} 
\end{equation}
where $c$ is the number of categories, $M(y_k)$ and $M(\hat{y}_k)$ denotes a set of elements of the binary masks of $y$  and $\hat{y}$ on class $k$, respectively.

In addition to the mean IoU accuracy, we also quantify statistics, i.e., the mean and standard deviation (std) of the results. Ideally, we expect to obtain a high mean value and a low std, i.e., the model can achieve equally accurate performance for different flight heights. 
 We consider a method to be a failure if it underperforms the trained model at the ground viewpoint, i.e, Ground-only.

\subsection{Quantitative Evaluation}
\subsubsection{Results on AirSim-Drone}
\label{sec_results_airsim}
The results of different approaches are given in Table~\ref{res_airsim_iou}. 
A close look at Ground-only shows a tendency of performance degradation from uav02 to uav10, and we observe a 45.7\% relative accuracy drop caused by the significant viewpoint difference. 

Moreover, both Pseudo-Labeling and ClassMix failed on the task, as they are even worse than Ground-only. 
Pseudo-Labeling demonstrates a lower accuracy with a mean of 0.457.
Although ClassMix is specially tailored to boost semantic segmentation, it significantly deteriorated due to the viewpoint difference.
We are not surprised that the previous methods of SSL failed. Intuitively, since clear viewpoint differences exist in data samples, they could not correctly predict labels from unlabeled data.

On the other hand, our method yields promising results for each sequence. Although we can also observe a tendency for performance degradation, 
the accuracy drop is only 20.7\% from uav02 to uav10. 
Compared with the original drop of 45.7\%, the improvement brought by our method is significant.
We obtained the mean of 0.599 and the std of 0.047, which shows 20.8\% and 55.2\% improvement from Ground-only of 0.496 and 0.105, respectively.

Fig.~\ref{fig_arisim} shows the estimated semantic maps provided by different methods at different flying heights. Clearly, ClassMix yielded inaccurate predictions, and it failed on all sequences. It tends to ignore small objects, and the estimated semantic maps have only three classes, i.e., Plant, Road, and Sky. When observing the results of Ground-only and Pseudo-Labeling, we can see how viewpoint differences mislead the semantic segmentation, e.g., in the left bottom areas of images, false positives gradually increase from 2 meters to 10 meters, while our method shows consistently correct results. 

Table~\ref{res_airsim_test} shows the generalization performance on the test sequence.
Although the accuracy is slightly lower than the results given in Table~\ref{res_airs_iou}, i.e., 0.599, our method achieved a 25.7\% performance boost from Ground-only, and other methods yielded worse results.

\subsubsection{Results on AIRs-Street}
\label{sec_results_airs}
The real-world AIRs-Street dataset is more challenging than the synthesized AirSim-Drone dataset, as it has more semantic attributes, few labeled images at the ground viewpoint, and dynamic objects, e.g., moving cars, and non-rigid objects, e.g., pedestrians, as well as different lighting conditions due to the change of the sunlight intensity. Thus, the performance of drone perception will also be deteriorated by the above difficulties other than the viewpoint difference.

Table~\ref{res_airs_iou} shows the numerical results for different methods. The results show that there is a 39.2\% accuracy drop of Ground-only from uav02 to uav09.
Also, ClassMix yielded inferior results for each sequence than Ground-only, and Pseudo-labeling obtained the same mean accuracy as Ground-only.
Our method outperformed Ground-only from uav02 to uav09. Similar to the results on the AirSim-Drone, the accuracy improvement tends to increase from lower flying height to higher flying height, e.g., there is only a 6.0\% boost for uav03 and a 32.2\% boost for uav09.
Overall, our method obtained a mean of 0.561 and a std of 0.003, which shows 16.9\% and 66.7\% performance improvement from Ground-only of 0.480 and 0.009, respectively.

\begin{figure*}[t]
\centering  
\begin{tabular}
{m{0.05\textwidth}<{\centering}m{0.13\textwidth}<{\centering}m{0.13\textwidth}<{\centering}m{0.13\textwidth}<{\centering}m{0.13\textwidth}<{\centering}m{0.13\textwidth}<{\centering}<{\centering}m{0.13\textwidth}<{\centering}} 
9m
&\IncG[width=0.1428\textwidth]{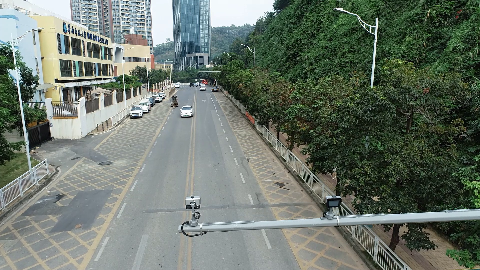}
&\IncG[width=0.1428\textwidth]{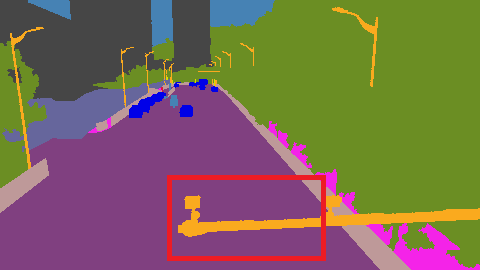}
&\IncG[width=0.1428\textwidth]{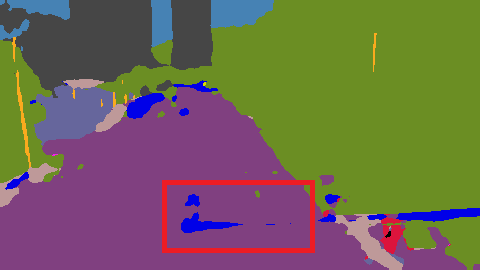}
&\IncG[width=0.1428\textwidth]{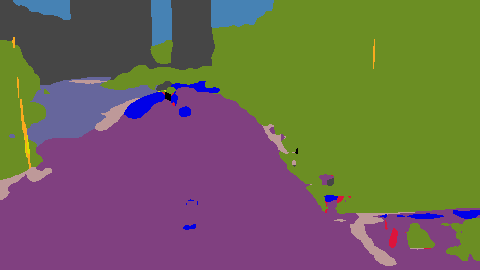}
&\IncG[width=0.1428\textwidth]{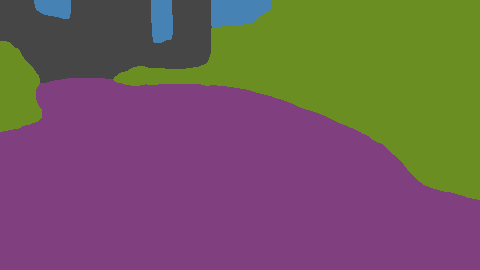}
&\IncG[width=0.1428\textwidth]{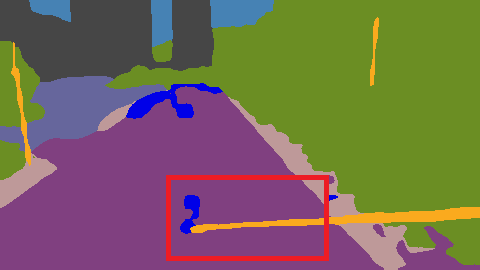}
\\
8m
&\IncG[width=0.1428\textwidth]{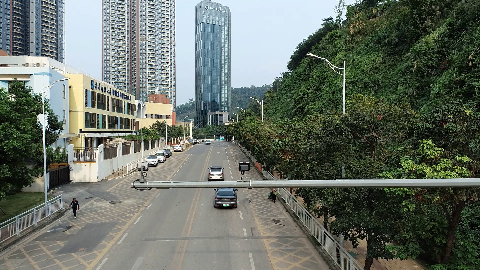}
&\IncG[width=0.1428\textwidth]{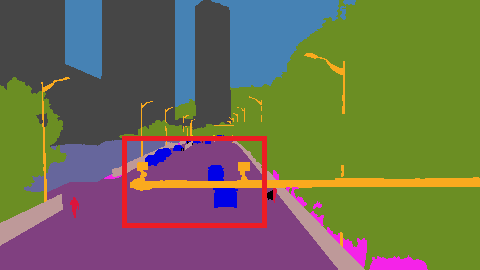}
&\IncG[width=0.1428\textwidth]{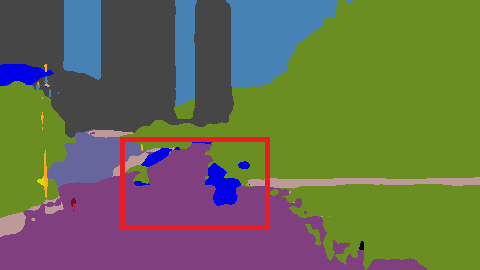}
&\IncG[width=0.1428\textwidth]{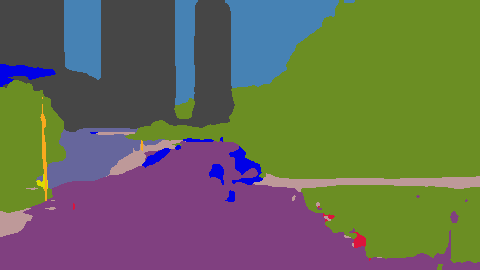}
&\IncG[width=0.1428\textwidth]{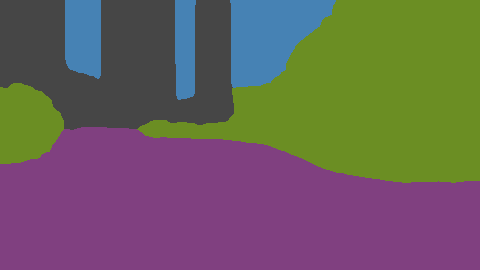}
&\IncG[width=0.1428\textwidth]{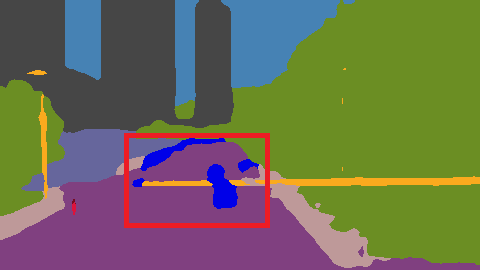}
\\
7m
&\IncG[width=0.1428\textwidth]{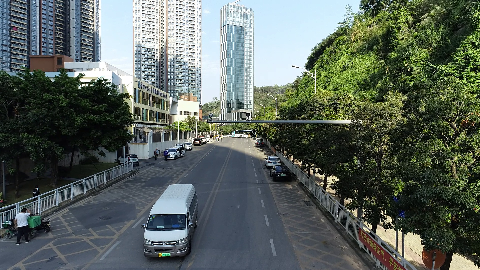}
&\IncG[width=0.1428\textwidth]{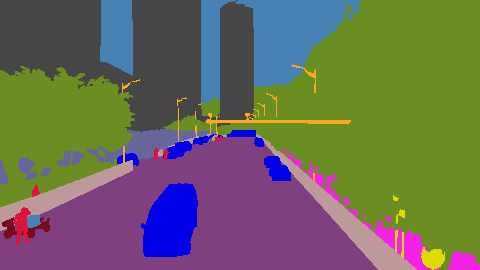}
&\IncG[width=0.1428\textwidth]{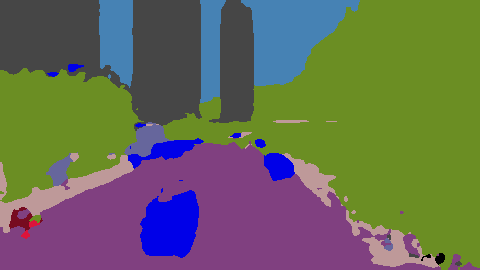}
&\IncG[width=0.1428\textwidth]{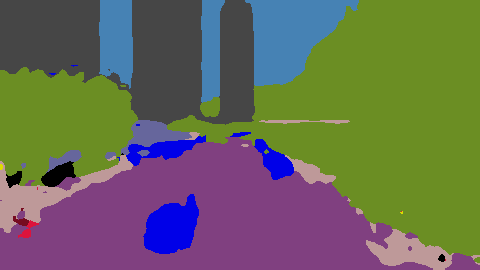}
&\IncG[width=0.1428\textwidth]{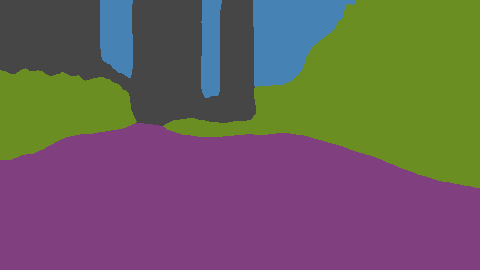}
&\IncG[width=0.1428\textwidth]{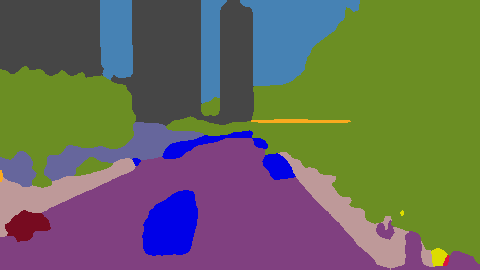}
\\

6m
&\IncG[width=0.1428\textwidth]{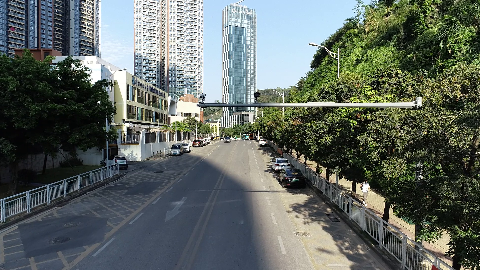}
&\IncG[width=0.1428\textwidth]{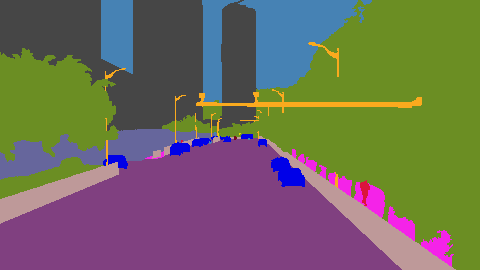}
&\IncG[width=0.1428\textwidth]{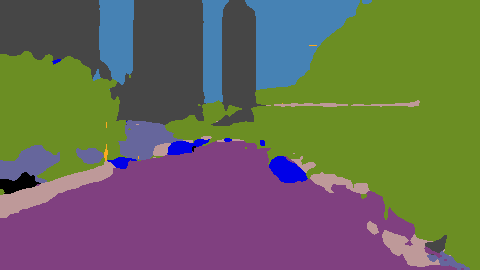}
&\IncG[width=0.1428\textwidth]{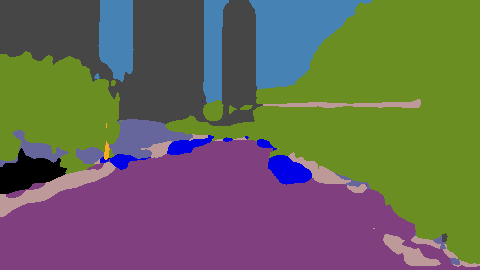}
&\IncG[width=0.1428\textwidth]{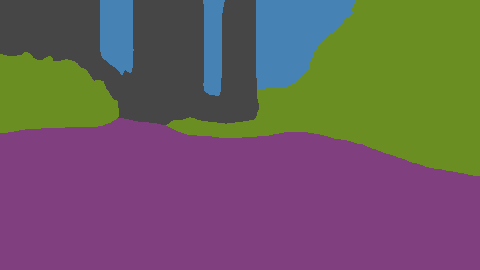}
&\IncG[width=0.1428\textwidth]{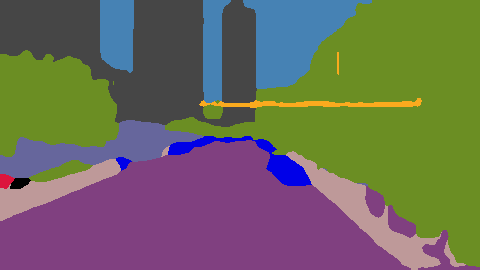}
\\
5m
&\IncG[width=0.1428\textwidth]{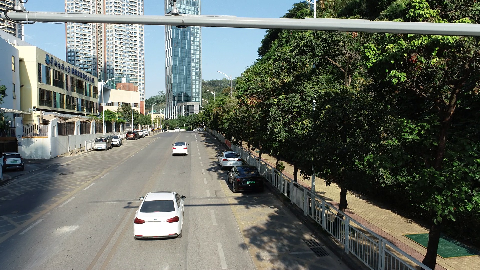}
&\IncG[width=0.1428\textwidth]{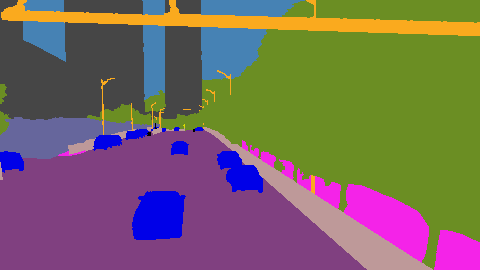}
&\IncG[width=0.1428\textwidth]{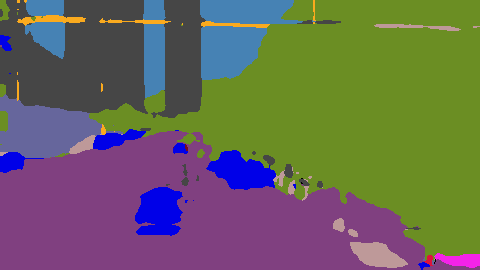}
&\IncG[width=0.1428\textwidth]{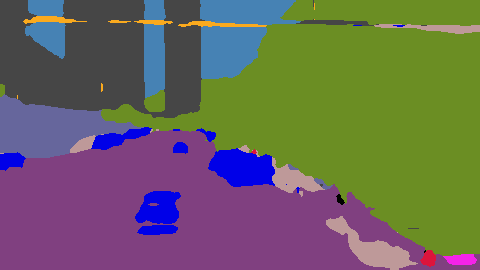}
&\IncG[width=0.1428\textwidth]{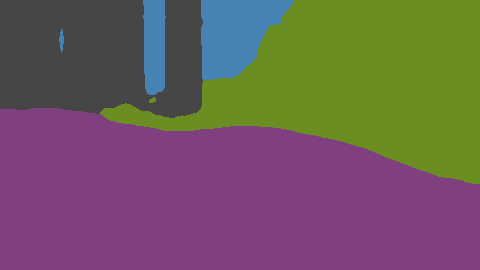}
&\IncG[width=0.1428\textwidth]{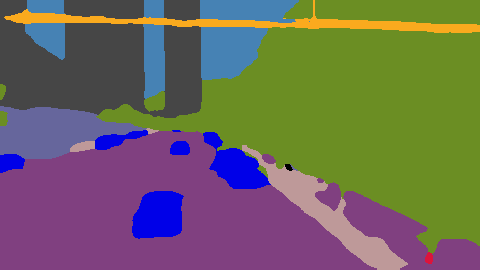}
\\
4m 
&\IncG[width=0.1428\textwidth]{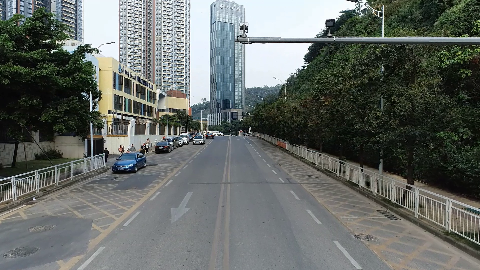}
&\IncG[width=0.1428\textwidth]{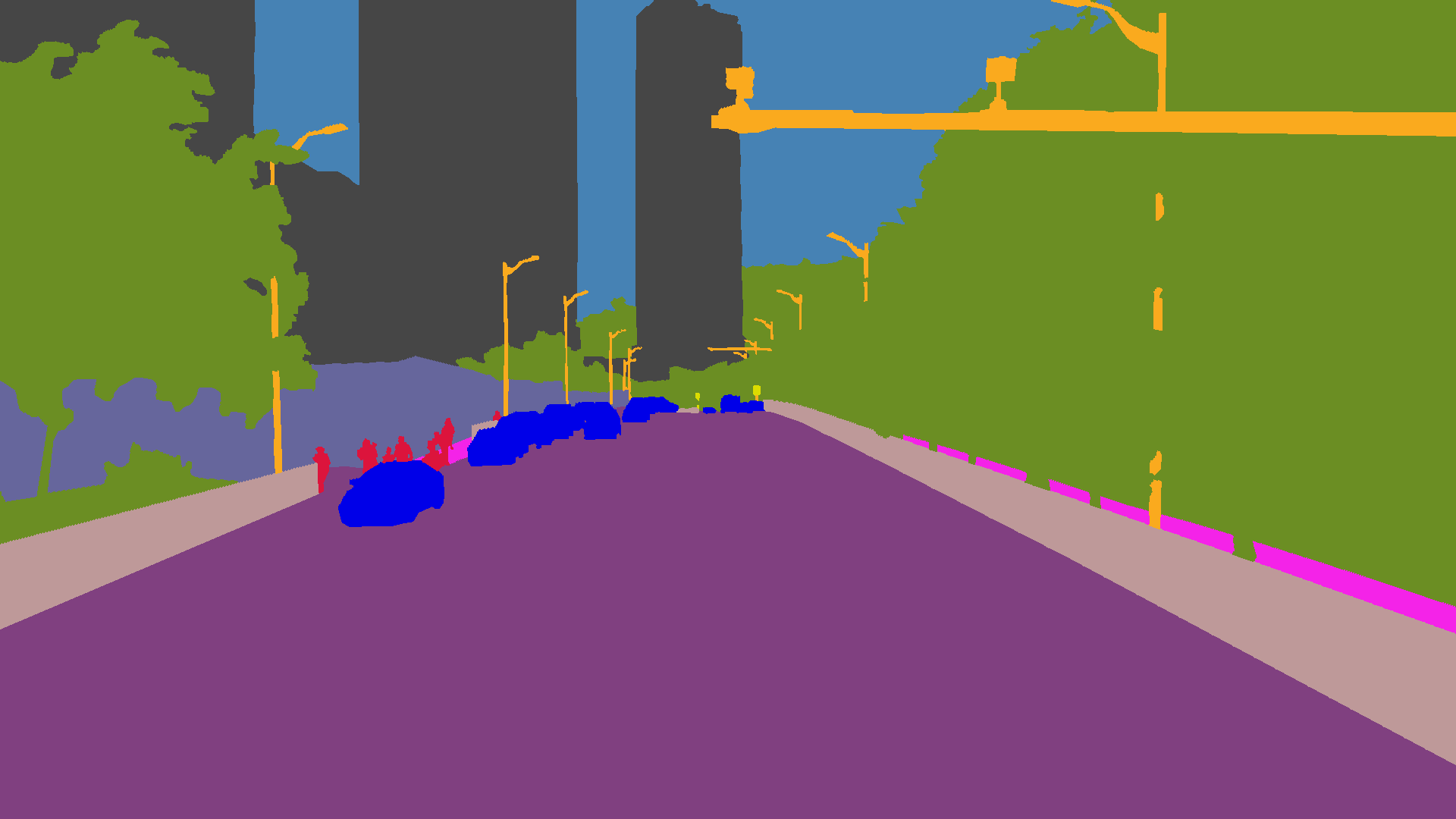}
&\IncG[width=0.1428\textwidth]{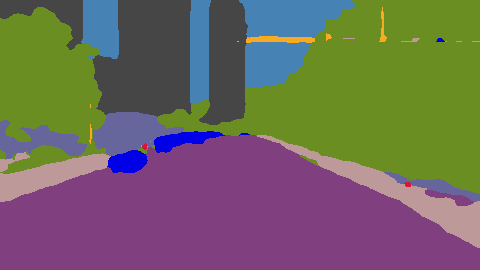}
&\IncG[width=0.1428\textwidth]{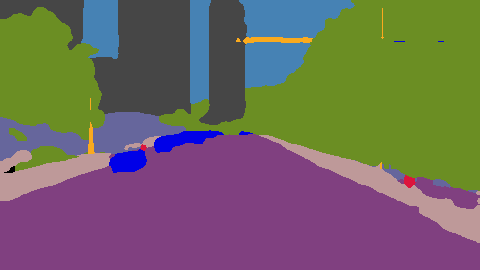} 
&\IncG[width=0.1428\textwidth]{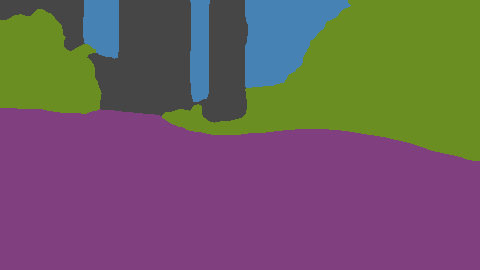}
&\IncG[width=0.1428\textwidth]{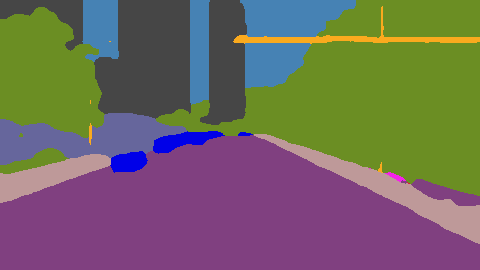}
\\
3m
&\IncG[width=0.1428\textwidth]{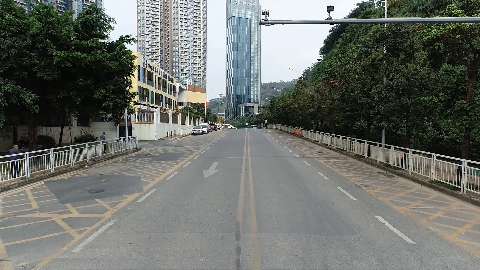}
&\IncG[width=0.1428\textwidth]{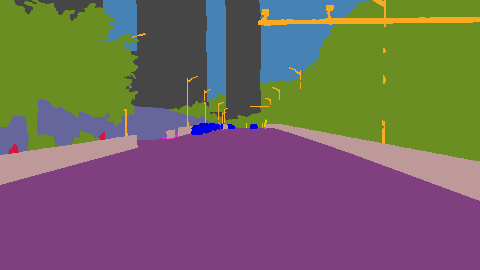}
&\IncG[width=0.1428\textwidth]{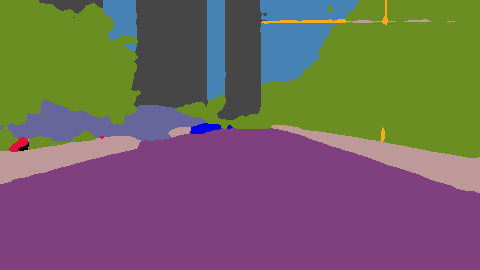}
&\IncG[width=0.1428\textwidth]{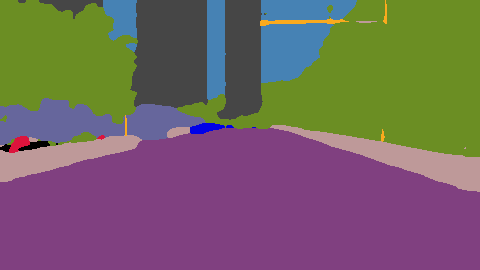}
&\IncG[width=0.1428\textwidth]{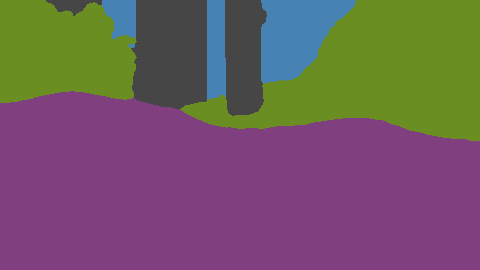}
&\IncG[width=0.1428\textwidth]{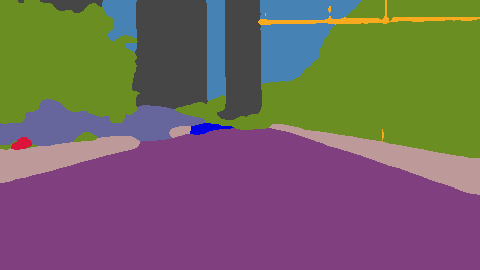}
\\
2m
&\IncG[width=0.1428\textwidth]{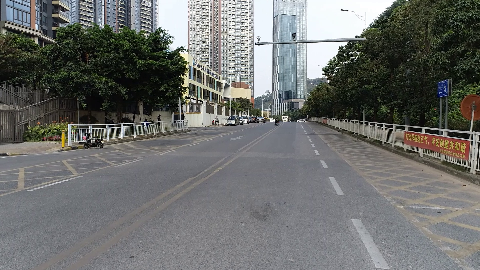}
&\IncG[width=0.1428\textwidth]{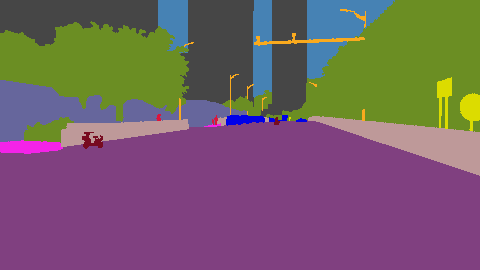}
&\IncG[width=0.1428\textwidth]{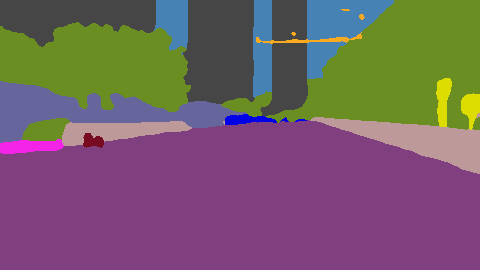}
&\IncG[width=0.1428\textwidth]{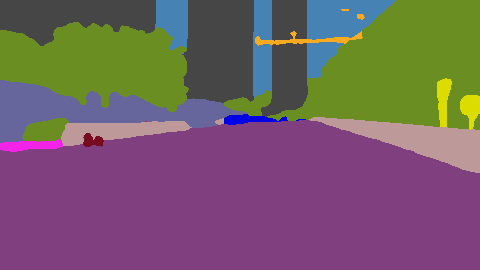}
&\IncG[width=0.1428\textwidth]{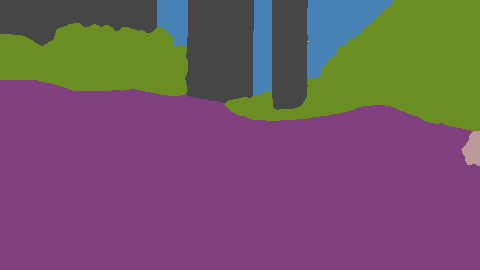}
&\IncG[width=0.1428\textwidth]{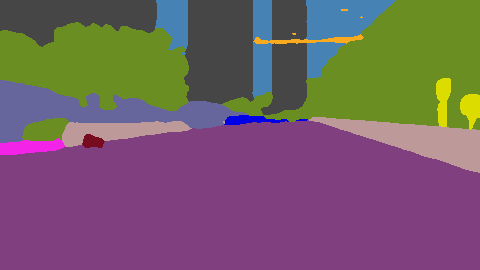}
\\
&\footnotesize{(a) Images}
& \footnotesize{(b) Ground Truths}
&\footnotesize{(c) Ground-only}
&{\fontsize{6.6}{7.92}\selectfont{(d) Pseudo-Labeling \cite{lee2013pseudo} }}
&\footnotesize{(e) ClassMix \cite{Olsson2021ClassMixSD}}
&\footnotesize{(f) Ours}
\\
\end{tabular}
\caption{Qualitative comparisons between our method and other approaches on the AIRs-Street dataset. We show the results for different flight heights from 2 meters to 9 meters. We draw red boxes on the Ground truth and results of Ground-only and Ours at 8m and 9m for better visualization.}
\label{fig_airs}
\end{figure*}

Fig.~\ref{fig_airs} shows the qualitative comparisons between our method and other approaches. The qualitative results agree well with those on AirSim-Drone.
ClassMix demonstrates inaccurate predictions for all flying viewpoints. Ground-only and Pseudo-Labeling gradually deteriorated from a flying height of 2 meters to 9 meters. As seen, false positives tend to increase for the two methods, and our method can correctly predict pixel attributes better, such as the cars.

\begin{table}[t]
\begin{center}
\caption{Generalization performance on the AirSim-Drone test set.}
\label{res_airsim_test}
\begin{tabular}
{r|l}
\hline
Methods &  Accuracy  \\ 
\hline
Ground-only &0.417  \\
Pseudo-Labeling \cite{lee2013pseudo}   &0.400   (-- 4.1\%)  \\
ClassMix \cite{Olsson2021ClassMixSD}    &0.179  (-- 57.1\%) \\ 
Ours     & \textbf{0.524}  (\textbf{+25.7\%}) \\
\hline
\end{tabular}
\end{center}
\end{table}

\begin{table*}[t]
\begin{center}
\caption{The mean IoU accuracy of our approach and other methods on the AIRs-Street dataset.}
\label{res_airs_iou}
\begin{tabular}
{m{0.15\textwidth}<{\raggedleft}|m{0.05\textwidth}<{\centering}m{0.05\textwidth}<{\centering}m{0.05\textwidth}<{\centering}m{0.05\textwidth}<{\centering}m{0.05\textwidth}<{\centering}m{0.05\textwidth}<{\centering}m{0.05\textwidth}<{\centering}m{0.05\textwidth}<{\centering}|m{0.03\textwidth}<{\centering}m{0.03\textwidth}<{\centering}}
\hline
Methods & uav02 & uav03 & uav04 & uav05 & uav06 & uav07 & uav08 & uav09  & mean & \hspace{1mm} std\\  \hline
Ground-only &0.607 &0.618 &0.563 & 0.458 &0.413 &0.408 &0.402 &0.369 & 0.480 & 0.009\\
Pseudo-Labeling \cite{lee2013pseudo}   &\textbf{0.611} &0.623 &0.546 &0.467  &0.406 &0.421 &0.386 &0.381 & 0.480 & 0.009 \\
ClassMix \cite{Olsson2021ClassMixSD}     &0.282 &0.260 &0.244 & 0.238 &0.233 &0.233 &0.229 &0.230&0.244 &\textbf{0.000} \\
Ours     &0.610 &\textbf{0.655} &\textbf{0.624} &\textbf{0.539} &\textbf{0.549} &\textbf{0.528} &\textbf{0.492}&\textbf{0.488} &\textbf{0.561} &0.003 \\
\hline
\end{tabular}
\end{center}
\end{table*}

\subsubsection{Adaptability to Various Flight Heights}
We set different height ranges starting from $h_2$ to maximum flight height $h_i$, in which $h_i$ ranges from $2,3,\ldots,10$ for AirSim-Drone and $2,3,\ldots,9$ for AIRs-Street, respectively. Then we evaluate our methods and baselines at these ranges.

\begin{figure}[t]
\centering
\subfigure[\footnotesize{Results on the AirSim-Drone.}]
{\includegraphics[width=0.24\textwidth]{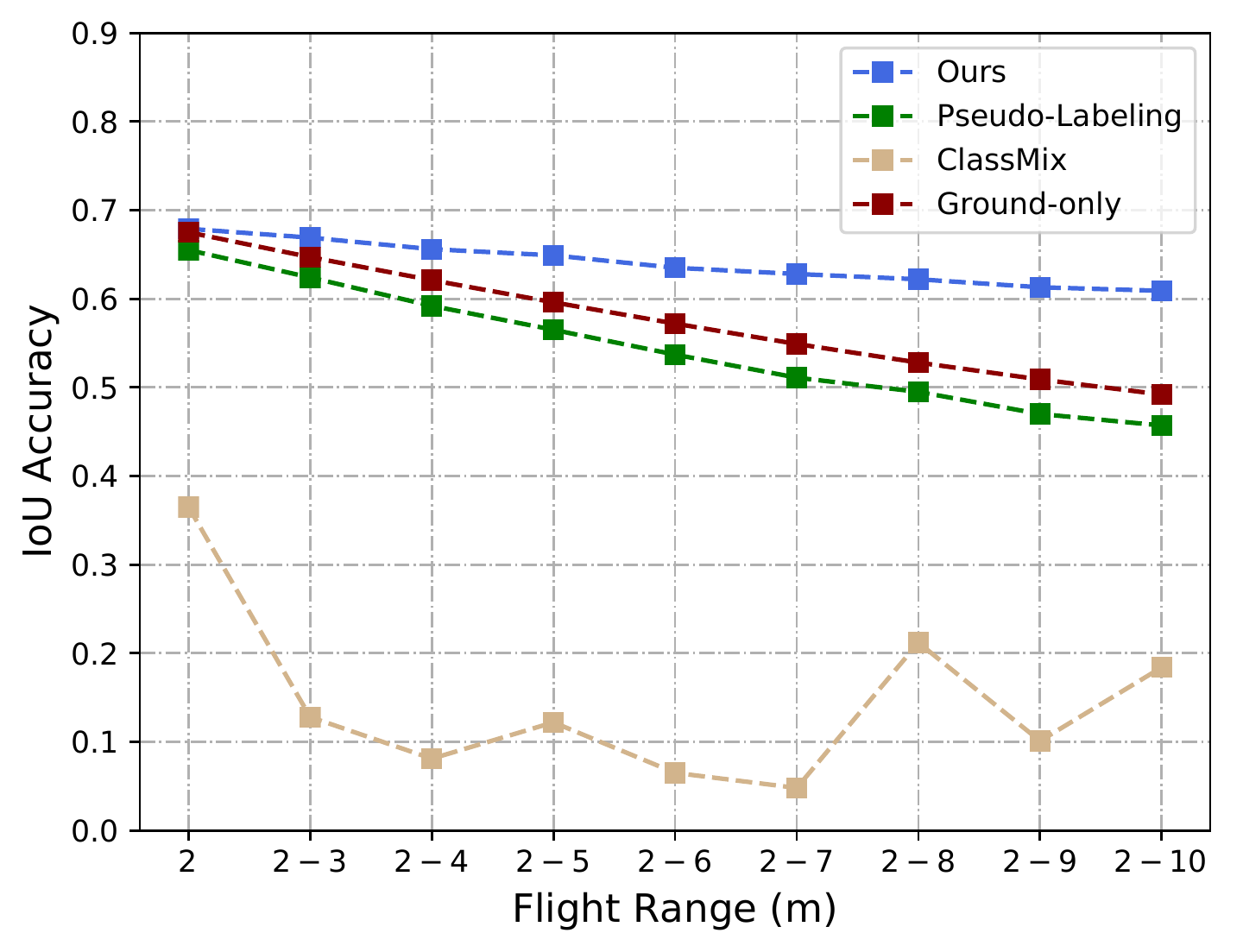}}
\subfigure[\footnotesize{Results on the AIRs-Street.}]
{\includegraphics[width=0.24\textwidth]{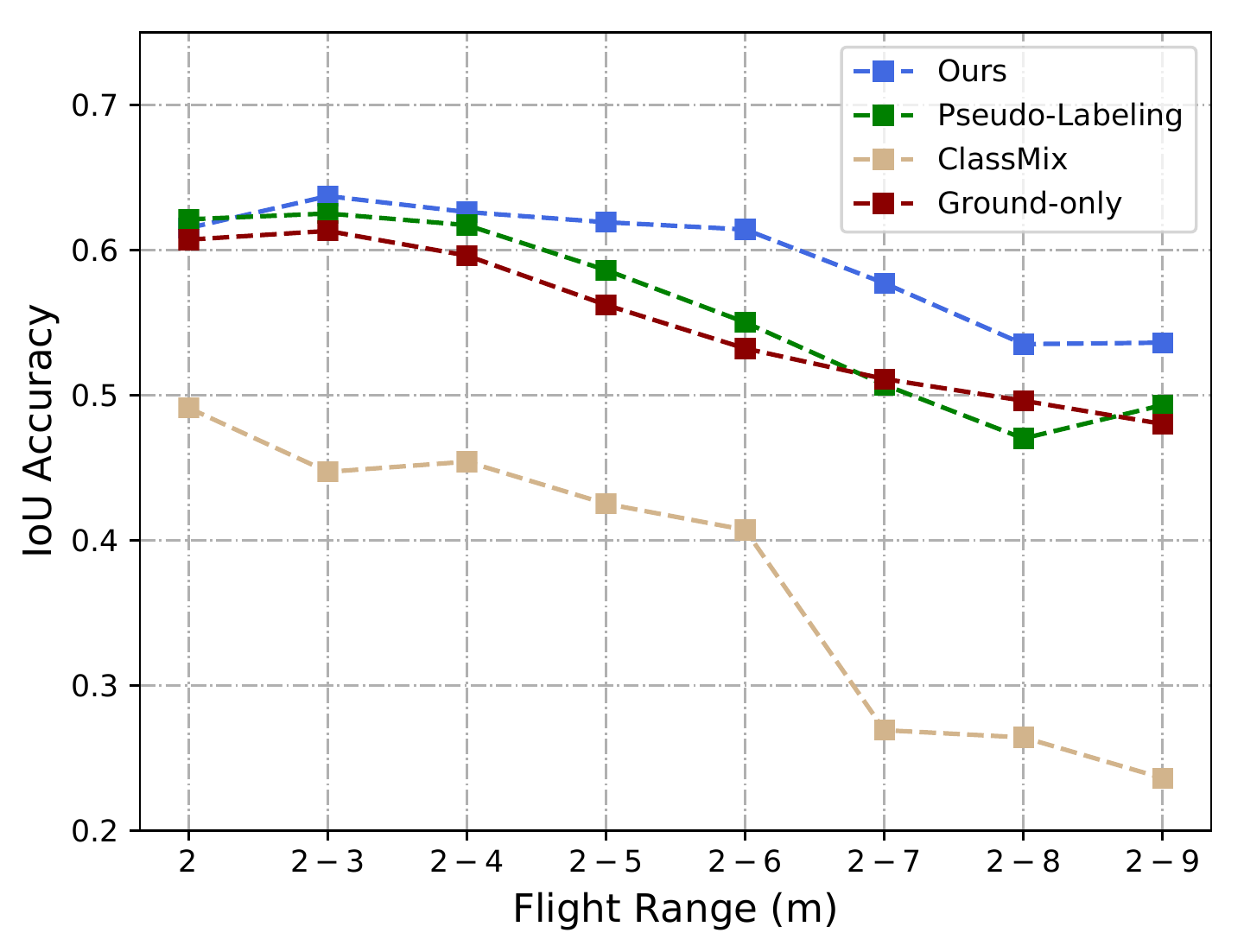}}
\vspace{-2mm}
\caption{Results for different methods while varying the maximum flying heights. The x-axis denotes the flight heights and the y-axis shows the mean accuracy on the corresponding range.}
\label{fig_maximum_heights}
\end{figure}

Fig.~\ref{fig_maximum_heights} shows the results on the two datasets. As seen, ClassMix demonstrated unstable behavior on AirSim-Drone and still had the worst performance on the two datasets.
Our method outperformed the other two approaches in all different settings, except for the maximum flight height of 2 meters on the AIRs-Street. 

\begin{figure}[t]
\centering
\subfigure[\footnotesize{Results on the AirSim-Drone.}]
{\includegraphics[width=0.24\textwidth]{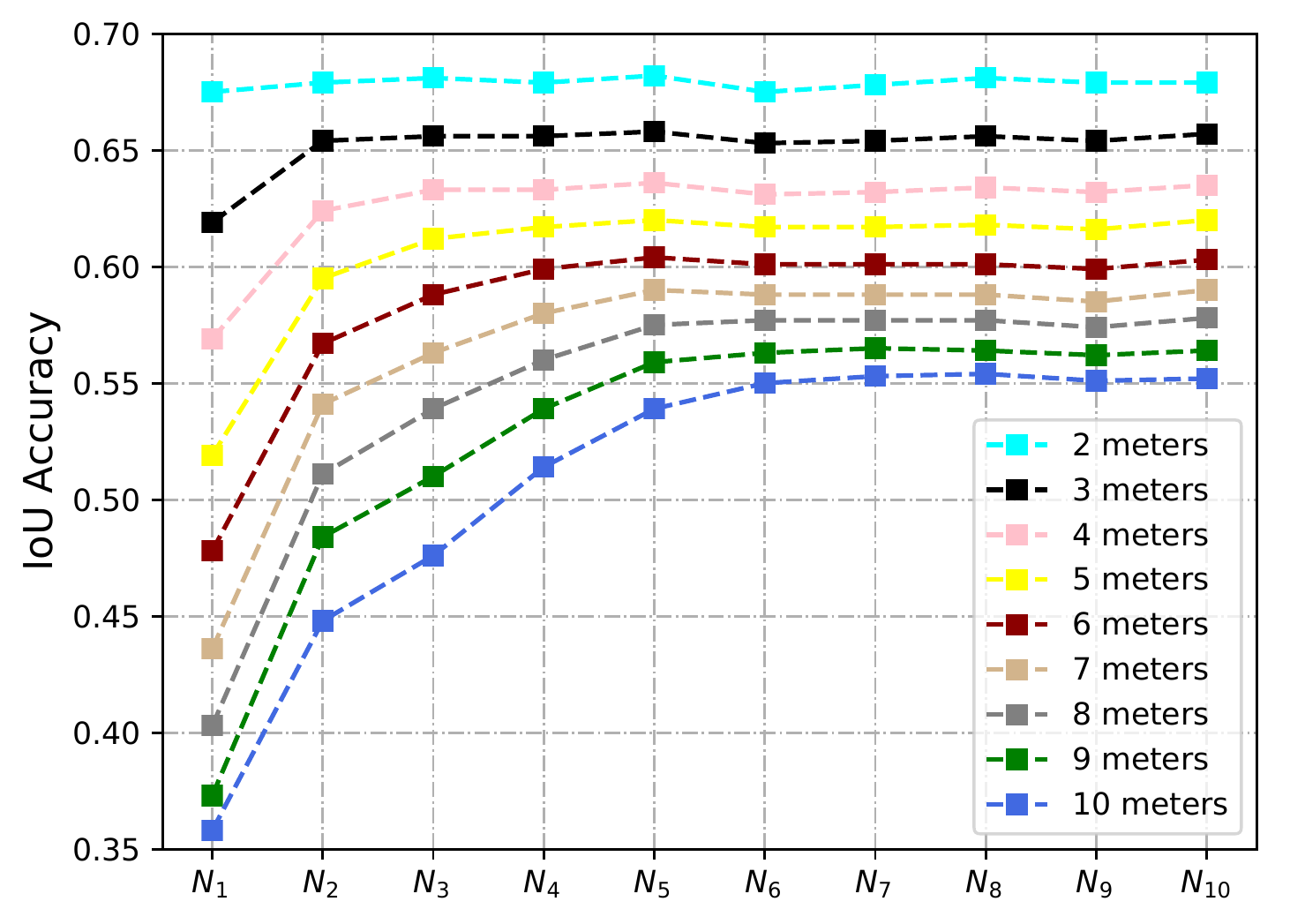}}
\subfigure[\footnotesize{Results on the AIRs-Street.}]
{\includegraphics[width=0.24\textwidth]{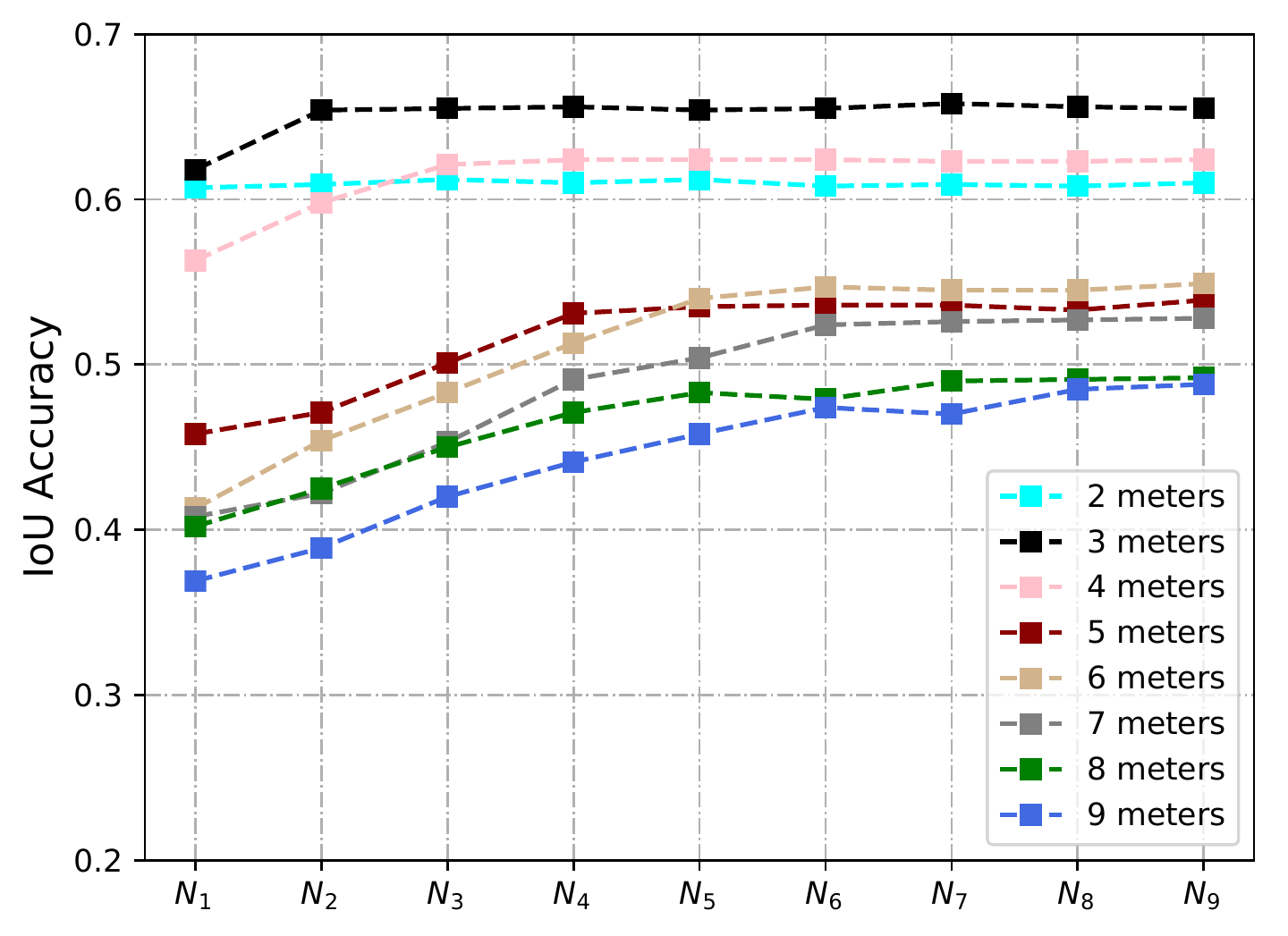}}
\vspace{-2mm}
\caption{Visualization of the application of our self-distillation method on each flight height. $N_i$ denotes the trained model for the maximum flight height $h_i$.}
\label{fig_iterations}
\end{figure}

We also visualize the accuracy change in our progressive learning framework for each flight height in Fig.~\ref{fig_iterations}, where $N_i$ denotes the trained model applied on the maximum flight height $h_i$, e.g., $N_5$ is learned from $h_1$ to $h_5$.
As seen in Fig.~\ref{fig_iterations} (a), our method demonstrates significant improvement for large viewpoint differences, e.g., uav10, while maintaining the accuracy for small viewpoint differences, e.g., uav02 on AirSim-Drone.
The same tendency can also be observed in Fig.~\ref{fig_iterations} (b).

Also, owing to the lack of supervision for flying viewpoints, we can observe that the increase in accuracy gradually slows down during the application of our self-distillation.
Since the viewpoint discrepancy increases gradually, we need additional supervision to handle some hard cases. We leave it for our future work.

\begin{table}[t]
\begin{center}
\caption{Split of AirSim-Drone with different sampling intervals.}
\label{sampling_rate}
\begin{tabular}
{m{0.055\textwidth}<{\centering}m{0.1\textwidth}<{\centering}m{0.25\textwidth}<{\centering}}
\hline
Intervals &Labeled view & Unlabeled view\\   \hline
1 meter & $h1$ & $h2,h3,h4,h5,h6,h7,h8,h9,h10$\\
2 meters &$h1$ & $h3,h5,h7,h9$\\
3 meters &$h1$ & $h4,h7,h10$\\
\hline
\end{tabular}
\end{center}
\end{table}

\begin{figure}[t]
\centering
\subfigure[Results for different sampling intervals.]{\includegraphics[width=0.32\textwidth]{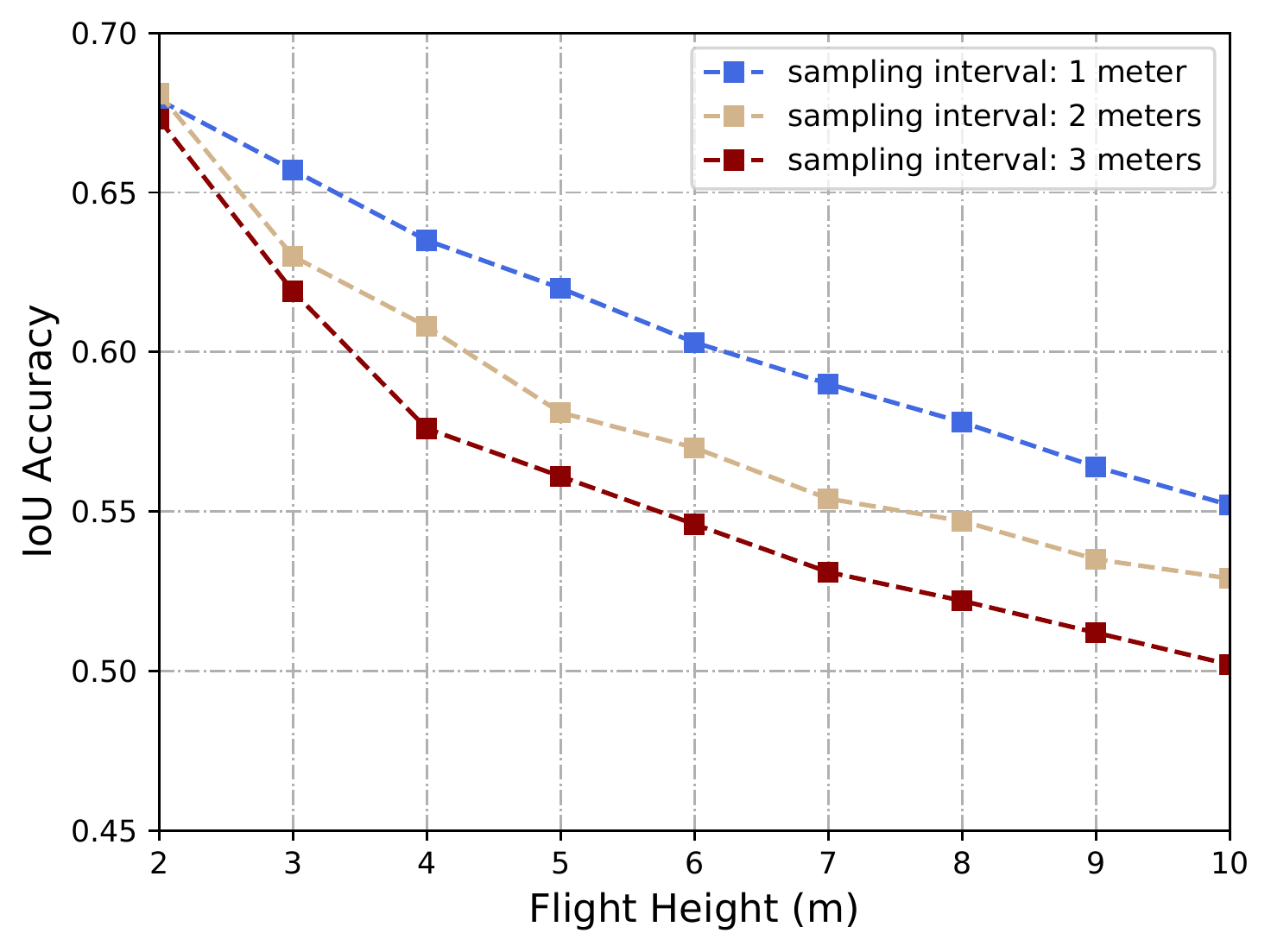}}
\subfigure[Results with and without using the MixView.]{\includegraphics[width=0.32\textwidth]{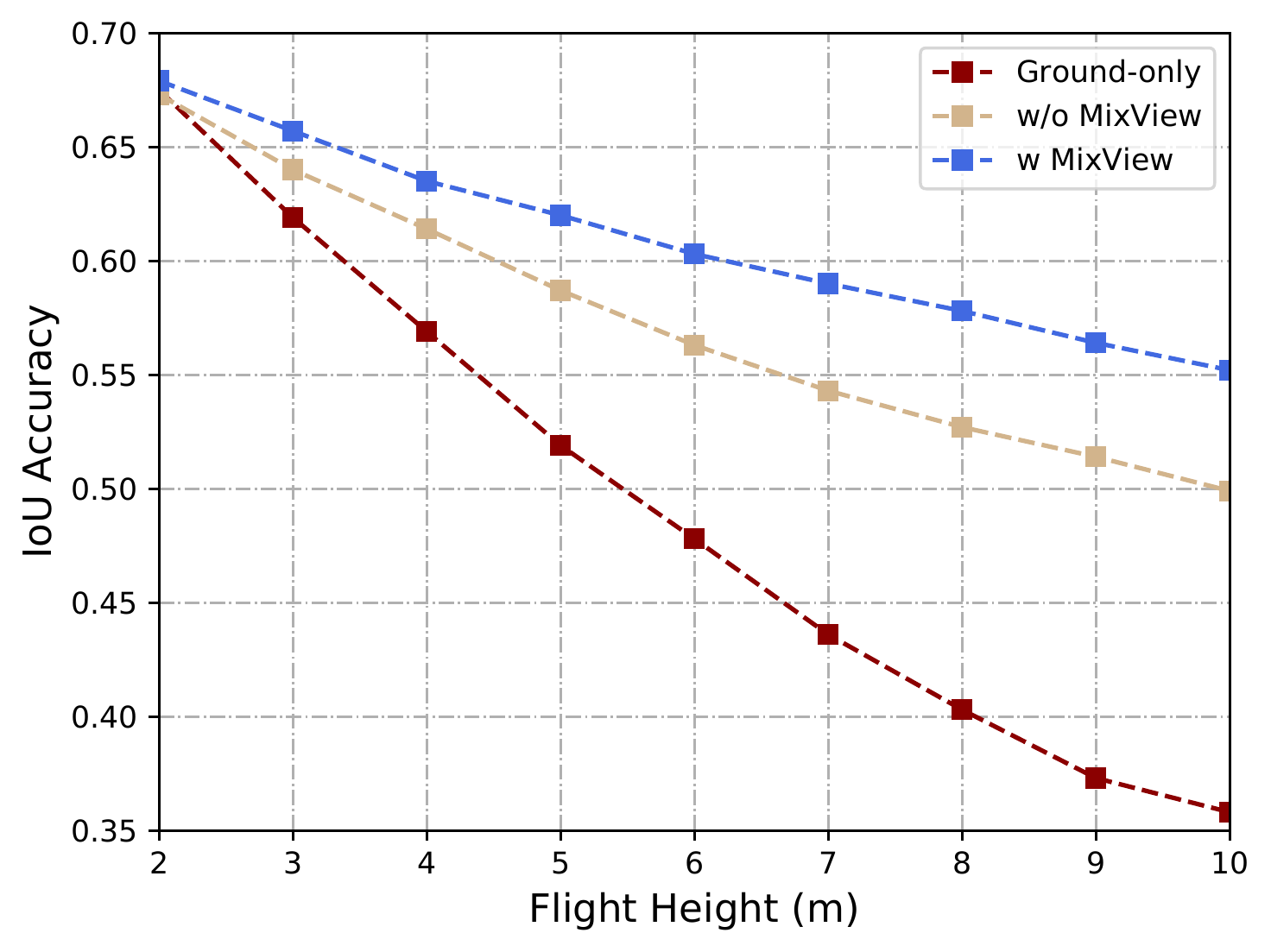}}
\subfigure[Results with and without using the nearest neighbor pseudo-labeling.]{\includegraphics[width=0.32\textwidth]{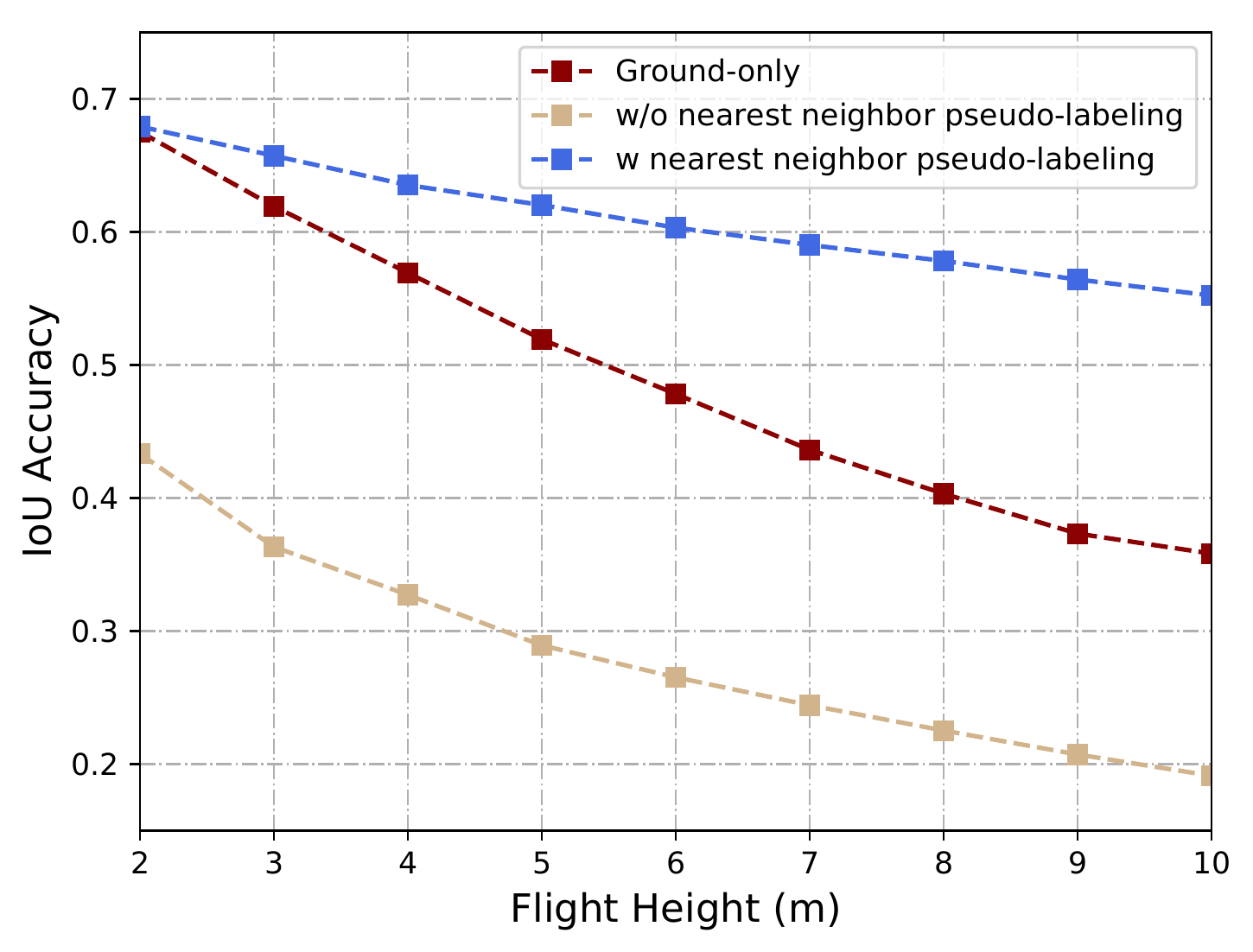}} 
\caption{Results of three ablation studies. We show the effect of different sampling intervals of viewpoints in (a), and performance with and without using MixView in (b), results with and without using the nearest neighbor pseudo-labeling in (c), respectively.}
\label{fig_res_ablation_study}
\end{figure}

\subsubsection{Summary}
Considering a model trained at the ground viewpoint as a baseline (i.e., a Ground-only model), we show through experiments on both a synthesized dataset and a real-world dataset that:
\begin{itemize}
\item The performance of a Ground-only model gradually deteriorates from lower flying height to higher flying height as viewpoint discrepancy gradually increases.

\item The viewpoint difference leads the previous SSL-based methods to malfunction. In our experiments, both Pseudo-Labeling and ClassMix failed on the task, as their performance is even worse than that of the Ground-only model. 

\item Our method shows a substantial performance boost compared to the Ground-only model. The mean relative accuracy improvement is 25.7\% and 16.9\% for the fixed AirSim-Drone and challenging AIRs-Street, respectively. 

\item Most importantly, we find that the performance improvement is significant for large viewpoint differences. We obtained 47.7\% and 32.2\% relative accuracy improvement for the flight height of 10 meters, i.e., uav10 of AirSim-Drone, and the flight height of 9 meters, i.e., uav09 of AIRs-Street, respectively.
 
\item Our method demonstrates good adaptability to various height ranges. It outperformed the other two SSL-based approaches in all different height ranges, except for the maximum flight height of 2 meters on the AIRs-Street.

\end{itemize}

\subsection{Ablation Studies}

We perform several ablation studies to analyze and understand better each component of our progressive SSL framework. To be specific, we conduct several experiments to investigate the 1) performance of different sampling intervals of viewpoints, 2) results with and without using the MixView, and 3) results with and without using the nearest neighbor pseudo-labeling. 
All ablation studies are performed on AirSim-Drone. The details are given as follows:
 
\paragraph{\textbf{Analyses with different sampling intervals}}
We evaluate the three settings of the sampling interval, i.e., 1 meter, 2 meters, and 3 meters. Detailed information on data sampling is given in Table~\ref{sampling_rate}. The experimental results are shown in Fig.~\ref{fig_res_ablation_study}. (a), where the blue, yellow, and red color denotes results for a sampling interval of 1, 2, and 3 meters, respectively. The results show that dense sampling of viewpoints contributes to better model performance, and we obtained 
$13.8\%$, $18.3\%$, and $23.8\%$ mean accuracy boost from the Ground-only model, respectively.

\paragraph{\textbf{Analyses with and without using the MixView}}
In our progressive learning framework, we propose to mix images of different viewpoints to generate augmented training samples. We provide experimental results to show that MixView can contribute to alleviating viewpoint differences. 
As seen in Fig.~\ref{fig_res_ablation_study}. (b), where red, yellow, and blue color denotes the results of the Ground-only model, our method $w/o$ and $w$ using the MixView, respectively. Even without utilizing MixView, we observe that the progressive distillation integrated with the nearest neighbor pseudo-labeling already outperformed Ground-only by $16.5\%$, and MixView can further improve the performance by $7.3\%$.

\paragraph{\textbf{Analyses with and without using the nearest neighbor pseudo-labeling}}
In our original setting, for training a model at $h_i$, we first use the trained model at $h_{i-1}$ to predict pseudo-labels from data captured at $h_i$. $N_i$ is then trained with pseudo-labeled samples of $h_1, \ldots, h_i$ and unlabeled samples of $h_i$.
We skip the above step to remove the effect of the nearest viewpoint pseudo-labeling, then the model $N_i$ is trained with pseudo-labeled samples of $h_1, \ldots, h_{i-1}$, and unlabeled samples of $h_i$. The results are shown in Fig.~\ref{fig_res_ablation_study}. (c), where red, yellow, and blue color denotes the results of the Ground-only model, our method $w/o$ and $w$ using the nearest neighbor pseudo-labeling, respectively. As a result, the mean accuracy degradation from the Ground-only model up to $47.8\%$ without utilizing the nearest neighbor pseudo-labeling.

\section{Conclusion}
In this paper, we have explored ground-to-aerial knowledge distillation to enable drone perception by learning from a UGV without increasing any flight data labeling cost. We argued that the essential requirement of drone perception is the capability to perceive from different flight heights, and the fundamental challenge is the significant viewpoint difference among flying heights.
Given only labeled images from the ground viewpoint and a set of unlabeled images from various flying viewpoints, we formulated it as a semi-supervised learning problem. We proposed a progressive learning framework that gradually learns from the ground viewpoint to the maximum flying height.
We proposed two methods to overcome the viewpoint difference. The first is the nearest neighbor pseudo-labeling that infers pseudo-labels of the nearest neighbor viewpoint with the model learned at the preceding viewpoint, and the second is the MixView which mixes data obtained from different viewpoints to generate viewpoint invariant data samples. 

To quantitatively and qualitatively verify the proposed method and inspire more future explorations, we created both a synthesized dataset and a real-world dataset. The experimental results show that the proposed method can yield promising results for different flight heights.
In practice, our approach can relax the requirement of data annotation for drone perception and expand a model for ground viewpoint perception to perform perception in fully 3D spaces. 
In the future, we will work on improving the training efficiency.
As a preliminary exploration on the task, we hope our work can inspire more future explorations in the community.

\bibliographystyle{IEEEtranS}
\bibliography{egbib}
\vspace{-11mm}
\begin{IEEEbiography}[{\includegraphics[width=0.8in,height=1in,clip,keepaspectratio]{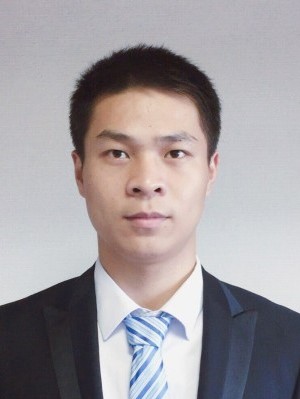}}]{Junjie Hu} (Member, IEEE) received the M.S. and Ph.D. degrees from the Graduate School of Information Science, Tohoku University, Sendai, Japan, in 2017 and 2020, respectively. He is currently a Research Scientist with the Shenzhen Institute of Artificial Intelligence and Robotics for Society. His research interests include artificial intelligence and robotics.
\end{IEEEbiography}
\vspace{-13mm}

\begin{IEEEbiography}[{\includegraphics[width=0.8in,height=1in,clip,keepaspectratio]{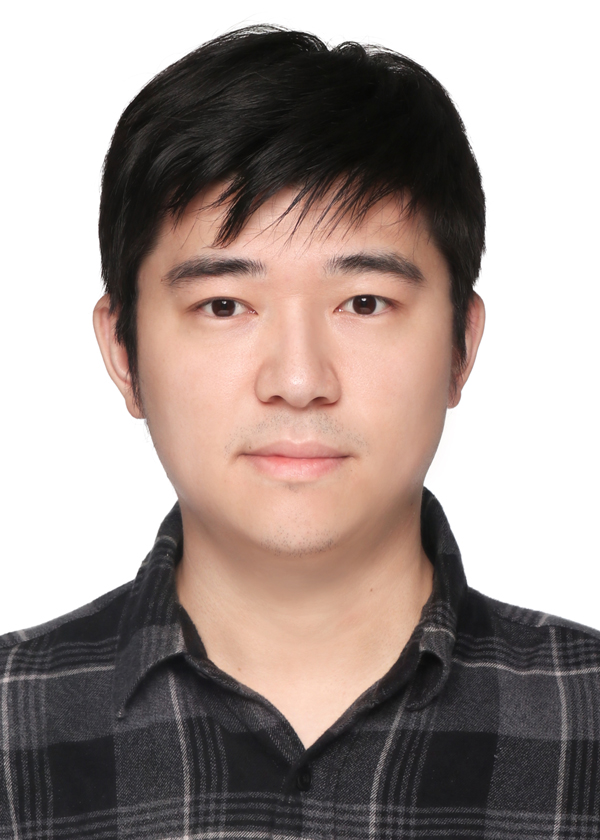}}]{Chenyou Fan} is an Associate Professor with the School of Artificial Intelligence, South China Normal University, China. 
He received the B.S. degree in computer science from the Nanjing University, China, in 2011, and the M.S. and Ph.D. degrees from Indiana University, USA, in 2014 and 2019, respectively. He was a Research Scientist with the Shenzhen Institute of Artificial Intelligence and Robotics for Society. His research interests include machine learning and computer vision.
\end{IEEEbiography}
\vspace{-13mm}

\begin{IEEEbiography}[{\includegraphics[width=0.8in,height=1in,clip,keepaspectratio]{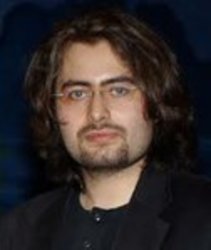}}]{Mete Ozay}  (M’09) received the B.Sc., M.Sc., Ph.D. degrees in mathematical physics, information systems, and computer engineering \& science from METU, Turkey. He has been a visiting Ph.D. and fellow in the Princeton University, USA, a research fellow in the University of Birmingham, UK, and an Assistant Professor in the Tohoku University, Japan. His current research interests include pure and applied mathematics,  theoretical computer science \& neuroscience.
\end{IEEEbiography}
\vspace{-13mm}

\begin{IEEEbiography}[{\includegraphics[width=0.8in,height=1in,clip,keepaspectratio]{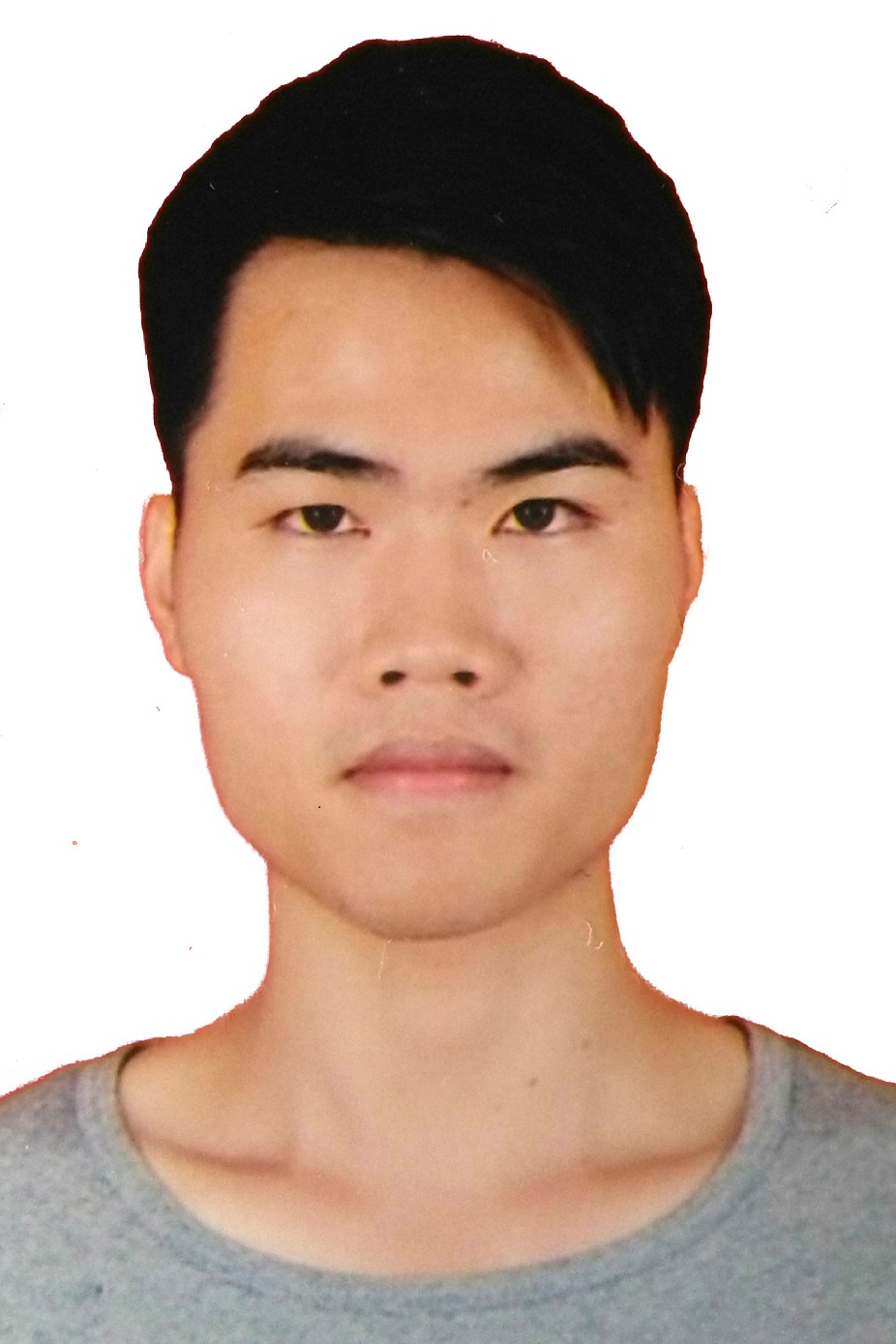}}]{Hua Feng} received the B.S. degree from the Dongguan University of Technology, Dongguan, China, in 2017, and master degree from the Wuyi University, Jiangmen, China, in 2022. He is currently a research intern with the Shenzhen Institute of Artificial Intelligence and Robotics for Society. His research interests include machine learning and robotics.
\end{IEEEbiography}
\vspace{-13mm}

\begin{IEEEbiography}[{\includegraphics[width=0.8in,height=1in,clip,keepaspectratio]{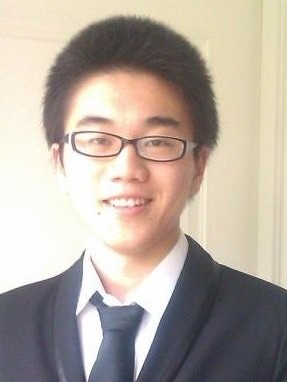}}]{Yuan Gao} received M.S. from the University of Helsinki in 2016 and Ph.D. degrees from the Department of Computer Science, Uppsala University, Uppsala, Sweden, in 2020, under the supervision of Prof. Ginevra Castellano and Prof. Danica Kragic. He is currently a Research Scientist with the Shenzhen Institute of Artificial Intelligence and Robotics for Society. His research interests include machine learning and robotics.
\end{IEEEbiography}
\vspace{-13mm}

\begin{IEEEbiography}[{\includegraphics[width=0.8in,height=1in,clip,keepaspectratio]{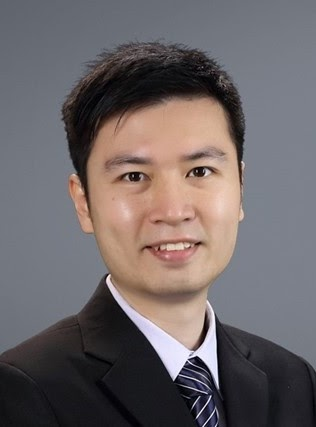}}]{Tin Lun Lam} (Senior Member, IEEE) received the Ph.D. degrees from the Chinese University of Hong Kong, Hong Kong, in 2010. 
He is an Assistant Professor with the Chinese University of Hong Kong, Shenzhen, China, and the Director of Center for the Intelligent Robots, Shenzhen Institute of Artificial Intelligence and Robotics for Society. He has published two monographs and more than 50 research papers in top-tier international journals and conference proceedings in robotics. His research interests include multi-robot systems, field robotics, and collaborative robotics. Dr. Lam received an IEEE/ASME T-MECH Best Paper Award in 2011 and the IEEE/RSJ IROS Best Paper Award on Robot Mechanisms and Design in 2020.
\end{IEEEbiography}

\appendix
\label{appe}
\paragraph{Definition of relative accuracy improvement}
Let Ground-only model be a baseline, we define the relative accuracy improvement as:
\begin{equation}
    RAI = \frac{ {Acc1-Acc2}}{ {Acc2}} 
    \label{RAI}
\end{equation}
Where Acc1 and Acc2 are IoU accuracy of our method and the ground-only model, respectively.
The accuracy improvements reported in experiments are calculated by Eq.\eqref{RAI}.

\paragraph{Difference between MixUp, ClassMix, and MixView}

MixUp and ClassMix are SSL techniques that utilize pixel-wise and object-wise mixing operations. Despite their effectiveness in certain scenarios, these methods fail to perform well under large viewpoint differences. To address this limitation, we propose MixView, an enhanced version of ClassMix specifically designed for handling viewpoint differences in SSL. This improvement is made possible by employing a dense sampling strategy to capture various flying heights and utilizing nearest neighbor pseudo-labeling to infer more accurate pseudo-labels.

\begin{table}[h]
\begin{center}
\caption{Comparison between MixUp, ClassMix, and MixView.}
\label{mix_dif}
\begin{tabular}
{lccc}
\hline
Method & Mixing operation &Accurate pseudo-labels & Viewpoint robust \\   \hline
MixUp & pixel-wise & \ding{55} & \ding{55} \\
ClassMix & object-wise & \ding{55} & \ding{55}   \\
MixView & object-wise & \ding{51} & \ding{51}  \\
\hline
\end{tabular}
\end{center}
\end{table}


\paragraph{Accuracy for each category }
To conduct more comprehensive evaluations,  we calculate the IoU accuracy for each category of the two datasets. Table~\ref{airsim_ps} shows the results on the Airsim-Drone dataset, indicating that our method demonstrates comparable performance on the ``Plant", ``Sky", and ``Ground"  categories, and superior results on the remaining six categories compared to the Ground-only method. Furthermore, on the AIRs-street dataset, our approach achieves marginally higher accuracy on the ``Plant", ``Road", 'Building' and ``Sky" categories, while exhibiting significant improvements on the remaining nine categories, as shown in Table~\ref{airs_ps}.

Upon examining these results and the proportion of pixels in each category, we can conclude that our method is particularly effective for large viewpoint differences. Specifically, objects such as ``Plant", ``Sky", and ``Road" that remain stationary and occupy a significant percentage of pixels in both datasets are more robust to viewpoint changes. These observations suggest that we can pay more attention to those small and dynamic objects for performance improvement in future exploration.


\begin{table}[t]
\begin{center}
\caption{IoU accuracy for each category on the AirSim-Drone test set.}
\renewcommand\arraystretch{1.2}
\label{airsim_ps}
\begin{tabular}
{r|c|cc}
\hline
 Categories & Percentage & Ground-only &Ours \\
\hline
Plant  &0.375 & \textbf{0.809}  & 0.805\\
 Sky  &0.210  &\textbf{0.885}  &0.879\\
 Road &0.186  & 0.630 & \textbf{0.727}\\
 Building &0.105  &0.563  & \textbf{0.780}\\
 Car  &0.010  &0.331  &\textbf{0.695}\\
 Ground &0.089 &\textbf{0.119} &0.117 \\
 Fence  &0.011  &0.186  &\textbf{0.333}\\
 Pole &0.002  &0.081  &\textbf{0.126}\\
Others &0.012  & 0.150 &\textbf{0.258}\\
\hline
\end{tabular}
\end{center}
\end{table}

\begin{table}[t]
\begin{center}
\caption{IoU accuracy for each category on the AIRs-Street (UAV02 to UAV09).}
\renewcommand\arraystretch{1.2}
\label{airs_ps}
\begin{tabular}
{r|c|cc}
\hline
 Categories & Percentage & Ground-only &Ours  \\
\hline
Plant  &0.326 &0.907  &\textbf{0.936} \\
Road &0.303 &0.902  &\textbf{0.931} \\
 Building &0.164  &0.915  &\textbf{0.933} \\
 Sky  &0.072 &0.942  &\textbf{0.947} \\
 Fence  &0.045 &0.465  &\textbf{0.641} \\
 Wall &0.034 &0.553  &\textbf{0.684} \\
 Car  &0.023 &0.655  &\textbf{0.766} \\
Sidewalk &0.015 &0.021  &\textbf{0.029} \\
Traffic Light  &0.009 &0.176  &\textbf{0.396}  \\
Person &0.006 &0.202  &\textbf{0.214} \\
Others &0.002 &0.062  &\textbf{0.100} \\
Motorcycle &0.001 &0.126  &\textbf{0.265} \\
Traffic Sign  &0.001 &0.299  &\textbf{0.408} \\
\hline
\end{tabular}
\end{center}
\end{table}

\end{document}